\documentclass[11pt]{article}
\usepackage[utf8]{inputenc}
\usepackage[preprint]{acl}
\usepackage{times}
\usepackage{latexsym}
\usepackage[T1]{fontenc}
\usepackage[utf8]{inputenc}
\usepackage{microtype}
\usepackage{inconsolata}
\usepackage{graphicx}
\usepackage{amsmath,amssymb}

\usepackage{algorithm}
\usepackage{algpseudocode}
\usepackage{booktabs}
\usepackage{tabularx}
\usepackage[normalem]{ulem} 
\usepackage[utf8]{inputenc}
\usepackage[normalem]{ulem}
\usepackage[table]{xcolor}
\usepackage{array}
\usepackage{graphicx}
\usepackage{subcaption} 
\newcommand{\pos}[1]{\textcolor{green!50!black}{\textnormal{#1}}}
\renewcommand{\neg}[1]{\textcolor{red!70!black}{\textnormal{#1}}}

\newcommand{\match}[1]{\colorbox{green!20}{#1}}    
\newcommand{\mismatch}[1]{\colorbox{red!15}{#1}}  
\newcommand{\improved}[1]{\colorbox{blue!15}{#1}} 

\title{Improving Low-Resource Machine Translation via Round-Trip Reinforcement Learning}

\author{
  Ahmed Attia \\
  MBZUAI \\
  Masdar City, UAE \\
  \texttt{ahmed.attia@mbzuai.ac.ae} \\\And
  Alham Fikri \\
  MBZUAI \\
  Masdar City, UAE \\
  \texttt{alham.fikri@mbzuai.ac.ae} 
   }

\usepackage{algorithmicx}
\begin{document}
\maketitle
\begin{figure*}[t]
  \centering
  \includegraphics[width=\textwidth]{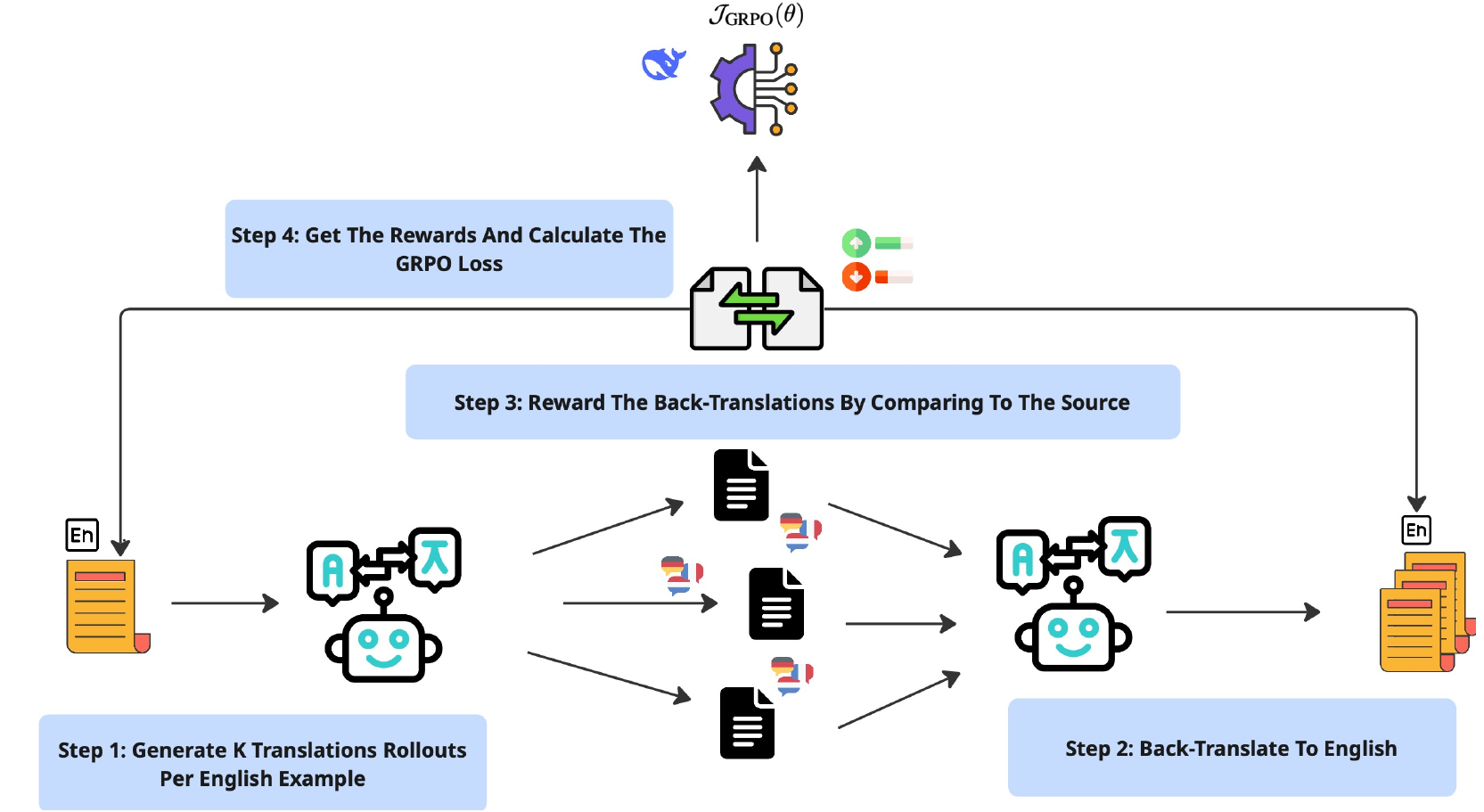}
  \caption{Overview Of Our Low-Resource Machine Translation With Round-Trip Reinforcement Learning Method.}
  \label{fig:method}
\end{figure*}

\begin{abstract}
Low-resource machine translation (MT) has gained increasing attention as parallel data from low-resource language communities is collected, but many approaches for improving low-resource MT remain underexplored. We investigate a self-supervised reinforcement learning fine-tuning for translation in low-resource settings using round-trip bootstrapping with the No Language Left Behind (NLLB) family of models. Our approach translates English into a target low-resource language and then back into English, using a combination of chrF++ and BLEU as the reward function on the reconstructed English sentences. Using the NLLB-MD dataset, we evaluate both the 600M and 1.3B parameter NLLB models and observe consistent improvements for the following languages: Central Aymara, Friulian, Wolof, Dyula, Bhojpuri and Russian. Qualitative inspection of translation outputs indicates increased fluency and semantic fidelity. We argue that our method can further benefit from scale, enabling models to increasingly leverage their pretrained knowledge and continue self-improving. Code available at: \url{https://github.com/Copticoder/MT-via-Round-Trip-RL}
\end{abstract}

\section{Introduction}

Low-resource machine translation remains one of the most persistent challenges in natural language processing. Modern neural translation systems achieve impressive performance for high-resource languages \citep{NIPS2017_3f5ee243, conneau-etal-2020-unsupervised, hendy2023good}, where millions of parallel sentence pairs are available for supervised training. However, thousands of languages lack parallel corpora of sufficient scale \citep{koehn2017six, haddow-etal-2022-survey}, leading to models that perform poorly or fail to generalize outside narrow training domains. As a result, speakers of many languages continue to face limited access to digital information and reduced participation in multilingual technologies.

Recent advances in massively multilingual machine translation have produced models that cover hundreds of languages within a single system. In particular, the No Language Left Behind (NLLB) project \cite{nllbteam2022languageleftbehindscaling} introduced a large-scale multilingual translation model trained on data covering more than 200 languages, including many low-resource ones. By learning shared multilingual representations, such models enable cross-lingual transfer that improves translation quality even when direct parallel supervision is limited. Nevertheless, adapting these models to further improve low-resource directions remains challenging when parallel data is unavailable.

In this work, we explore whether pretrained multilingual translation models can improve translation quality using only monolingual signals. We introduce a self-supervised framework that combines round-trip translation with reinforcement learning (RL). Given an English sentence, the model first generates a translation into a target low-resource language and then translates the result back into English. If the translation preserves the original meaning, the reconstructed sentence should closely match the source. We therefore use similarity metrics such as BLEU and chrF++ to provide rewards that guide learning.

Standard machine translation training relies on maximum likelihood estimation (MLE) \citep{Fisher1992}, which predicts tokens conditioned on ground-truth histories. While effective, MLE suffers from exposure bias \citep{bengio2015scheduled, wang2020exposure, he-etal-2024-recovery} and an objective mismatch \citep{maksai2018eliminatingexposurebiaslossevaluation}, since training does not directly optimize evaluation metrics such as chrF++ \citep{popovic2015chrf}. These issues are particularly pronounced in low-resource settings, where limited supervision makes models more susceptible to compounding errors during inference.

Back-translation \citep{sennrich2016improvingneuralmachinetranslation} is the dominant approach for leveraging monolingual data in low-resource MT. However, its effectiveness depends heavily on the quality of the reverse translation model, and errors in early translations can propagate through synthetic training data \citep{mcnamee2023extensive}. Moreover, standard self-supervised approaches typically rely on maximum likelihood training rather than directly optimizing translation metrics.

Reinforcement learning provides a natural framework for addressing these limitations. By treating translation as a sequential decision-making process, RL allows models to optimize non-differentiable evaluation metrics and learn from their own predictions during training \citep{Sutton1998}. Prior work has explored RL for neural machine translation \citep{ranzato2016sequenceleveltrainingrecurrent, wu2018studyreinforcementlearningneural}, and recent studies have applied RL techniques to large language models for translation \citep{feng2025mtr1zeroadvancingllmbasedmachine}. However, the use of RL to improve round-trip translation consistency for low-resource languages remains relatively underexplored.

We propose a reinforcement learning approach that directly optimizes round-trip reconstruction quality using Group Relative Policy Optimization (GRPO) \cite{shao2024deepseekmathpushinglimitsmathematical} applied to pretrained NLLB models \cite{nllbteam2022languageleftbehindscaling}. This approach allows the model to improve translation quality without requiring additional parallel data.

Our contributions are as follows:

\begin{itemize}
\item We introduce a reinforcement learning framework for improving low-resource translation using round-trip consistency without parallel training data.

\item We provide a systematic analysis of reward functions, showing that chrF++ is a more effective training signal than BLEU or learned metrics.
\item We provide empirical analysis across several languages and model sizes.
\end{itemize}

\section{Related Work}

\subsection{Improving NMT with Monolingual Data and Round-Trip Translations}

Back-translation is a widely used technique for leveraging monolingual data in neural machine translation. \citet{sennrich2016improvingneuralmachinetranslation} showed that generating synthetic source sentences from target-language monolingual data can significantly improve translation quality when combined with genuine parallel corpora.

Building on this idea, unsupervised machine translation methods have demonstrated that translation systems can be trained without parallel data by combining denoising autoencoding, back-translation, and cycle-consistency objectives \citep{lample2018unsupervisedmachinetranslationusing}. These approaches enforce consistency between forward and backward translation models, encouraging translations that preserve the original meaning after a round-trip process.

Round-trip translation has also been explored in other contexts. Early work by \citet{huang-1990-machine} used round-trip translation as a quality-checking tool for users unfamiliar with the target language. \citet{federmann-etal-2019-multilingual} later demonstrated that translating through multiple pivot languages can generate diverse paraphrases, a process known as ``multilingual whispers.''

Other self-supervised approaches attempt to mine parallel sentences from comparable corpora. For example, \citet{ruiter-etal-2019-self} propose jointly mining sentence pairs and training translation models using cross-lingual similarity signals derived from internal model representations. Their results show that high-quality translation models can be trained from comparable data without direct parallel supervision.

\subsection{Reinforcement Learning for Neural Machine Translation}

Reinforcement learning has been explored as a way to optimize sequence-level objectives in machine translation. \citet{ranzato2016sequenceleveltrainingrecurrent} and \citet{wu2018studyreinforcementlearningneural} showed that policy gradient methods can reduce exposure bias and directly optimize evaluation metrics. These approaches typically combine RL objectives with maximum likelihood training to stabilize optimization.

More recently, RL techniques have been applied to large language models for translation. \citet{feng2025mtr1zeroadvancingllmbasedmachine} propose MT-R1-Zero, which applies RL without supervised warm-starts and uses rule-based reward functions to improve translation quality. Their method employs Group Relative Policy Optimization (GRPO) \cite{shao2024deepseekmathpushinglimitsmathematical}, an algorithm derived from Proximal Policy Optimization (PPO) \citep{schulman2017proximalpolicyoptimizationalgorithms} that removes the value critic and normalizes rewards across sampled outputs.
\section{Method}
We propose a reinforcement learning framework that leverages round-trip translation consistency to fine-tune pretrained multilingual language models for low-resource machine translation. Our approach combines Group Relative Policy Optimization (GRPO) \cite{shao2024deepseekmathpushinglimitsmathematical} with back-translation-based rewards, enabling effective adaptation without parallel data augmentation.

\subsection{Preliminary}

For each input $q$, GRPO \cite{shao2024deepseekmathpushinglimitsmathematical} samples a group of $G$ candidate outputs
$\{o_1,\dots,o_G\} \sim \pi_{\theta_{\text{old}}}(y \mid x)$ from the
current policy. The update maximizes
\small
\begin{equation}
\begin{aligned}
\mathcal{J}_{\text{GRPO}}(\theta)
&= \mathbb{E}_{q\sim P(Q),\,\{o_i\}_{i=1}^G \sim \pi_{\theta_{\text{old}}}(O\mid q)}
\Bigg[ \\
&\quad
\frac{1}{G}\sum_{i=1}^G \frac{1}{|o_i|}\sum_{t=1}^{|o_i|}
\min\!\Big( r_{i,t}(\theta)\,\hat A_{i,t}, \\
&\qquad\qquad
\operatorname{clip}\big(r_{i,t}(\theta),1-\varepsilon_c,1+\varepsilon_c\big)\,
\hat A_{i,t} \Big) \\
&\quad
- \beta\, D_{\mathrm{KL}}\!\big[\pi_\theta \Vert \pi_{\mathrm{ref}}\big]
\Bigg], \\
&\text{where } r_{i,t}(\theta)
= \frac{\pi_\theta(o_{i,t}\mid q,o_{i,<t})}
       {\pi_{\theta_{\text{old}}}(o_{i,t}\mid q,o_{i,<t})}.
\end{aligned}
\end{equation}
\normalsize
GRPO differs from PPO in how it defines the advantages $\hat A_{i,t}$.
Instead of subtracting a critic prediction, GRPO normalizes rewards
within the sampled group. Let $R_i = R(x,y_i)$; then
\begin{equation}
\hat A_{i,t} =
\frac{R_i - \operatorname{mean}(\{R_1,\dots,R_G\})}
     {\operatorname{std}(\{R_1,\dots,R_G\})}.
\end{equation}

When all or none of the candidates solve the problem, every $\hat A_{i,t}$ is
zero and the policy gradient vanishes except for the KL term.

\subsection{Problem Formulation}

We consider the setting where \emph{no parallel corpus is available} for the source and target language pair. Instead, we assume access to a pretrained multilingual neural machine translation (NMT) model $\pi_\theta$ and monolingual corpora in the source language (English). Our goal is to improve the translation quality in both directions for this low-resource pair \emph{without} relying on any parallel data for training.

We formulate this adaptation problem as a reinforcement learning problem. At each step, the model samples translation candidates and receives rewards that reflect translation quality, defined purely from intrinsic signals. Concretely, starting from a source sentence $x$, the model generates a target hypothesis $\tilde{y} \sim \pi_\theta(\cdot \mid x)$, then translates back $\hat{x} \sim \pi_\theta(\cdot \mid \tilde{y})$. The agent is rewarded according to the quality of the round-trip reconstruction, for example via a similarity measure (chrF++) between $x$ and $\hat{x}$. In this way, the model learns from self-consistency and monolingual structure, without any parallel supervision.

Let $x$ denote a source sentence and $y$ denote its target translation. The policy $\pi_\theta(y|x)$ produces candidate translations by autoregressively sampling tokens $y = (y_1, y_2, \ldots, y_T)$ conditioned on the source.

\subsection{Round-Trip Translation Reinforcement Training}

Our key insight is that translation consistency in both directions provides a stronger learning signal than unidirectional training alone. For each source sentence $x$, we execute a two-phase generation process:

\paragraph{Phase 1: Forward Translation.} Generate $K$ candidate translations from source to target:
\begin{equation}
    \hat{y}^{(k)} \sim \pi_\theta(\cdot \mid x), \quad k \in \{1, \ldots, K\}
\end{equation}

\paragraph{Phase 2: Back-Translation.} For each forward translation $\hat{y}^{(k)}$, generate $K$ back-translations to the source language:
\begin{equation}
    \hat{x}^{(k,j)} \sim \pi_\theta(\cdot \mid \hat{y}^{(k)}), \quad j \in \{1, \ldots, K\}
\end{equation}

The model is trained on the back-translation phase, where rewards measure how well the back-translated sentence $\hat{x}$ matches the original source $x$. This self-consistency objective encourages the model to produce translations that preserve semantic content bidirectionally.

\subsection{Reward Design}

We employ a character-level reward function using chrF++.

\paragraph{chrF++.} \cite{popovic2015chrf} is a character n-gram F-score that is particularly suitable for morphologically rich and low-resource languages, as it provides smoother gradients than word-based metrics:
\begin{equation}
    \text{chrF++} = (1 + \beta^2) \cdot \frac{\text{chrP} \cdot \text{chrR}}{\beta^2 \cdot \text{chrP} + \text{chrR}}
\end{equation}

\paragraph{BLEU.} \cite{papineni-etal-2002-bleu} provides complementary word-level precision signals. We mainly use BLEU in our ablation studies.


\begin{table*}[!t]
\centering
\small
\setlength{\tabcolsep}{5.5pt}
\renewcommand{\arraystretch}{1.15}
\caption{Main results (higher is better). For each language and model size, we \textbf{bold} the best score and \underline{underline} the second-best across \{Base, BT, UMT, RT, Ours\}. BT denotes target-to-source back-translation (tgt$\rightarrow$src). UMT (Unsupervised Machine Translation with Monolingual Corpora Only), RT (round-trip), and Ours are trained via round-trip translation (src$\rightarrow$tgt$\rightarrow$src).}
\label{tab:main_results_roundtrip_rl}
\begin{tabular}{lcccccccccc}
\toprule
& \multicolumn{5}{c}{NLLB-600M} & \multicolumn{5}{c}{NLLB-1.3B} \\
\cmidrule(lr){2-6}\cmidrule(lr){7-11}
Language & Base & BT & UMT & RT & Ours & Base & BT & UMT & RT & Ours \\
\midrule
Central Aymara (ayr\_Latn) & 24.38 & \underline{27.34} & 24.85 & 26.68 & \textbf{28.13} & 26.01 & \underline{28.61} & 28.21 & 27.59 & \textbf{28.73} \\
Wolof (wol\_Latn)          & 20.73 & \underline{23.57} & 19.89 & 22.84 & \textbf{23.96} & 24.73 & 26.35 & \textbf{27.72} & 25.21 & \underline{26.66} \\
Russian (rus\_Cyrl)        & 48.80 & \textbf{53.31} & 51.70 & 51.63 & \underline{52.79} & 50.10 & \textbf{55.46} & 55.14 & 54.07 & \underline{55.16} \\
Friulian (fur\_Latn)       & 45.75 & 47.38 & \textbf{48.96} & 46.25 & \underline{47.61} & 49.41 & \textbf{51.72} & 46.43 & 49.68 & \underline{50.76} \\
Dyula (dyu\_Latn)          & 18.03   & 20.13 & 20.82 & \underline{21.89} & \textbf{22.50} & 18.24   & 20.89 & \underline{22.93} & 21.96 & \textbf{22.97} \\
Bhojpuri (bho\_Deva)       & 43.07   & \textbf{45.36} & 39.79 & 41.62 & \underline{43.97} & 44.31   & \textbf{47.23} & 44.43 & 44.52 & \underline{44.77} \\
\midrule
Avg. & 34.91 & 36.18 & 34.34 & 35.15 & \textbf{36.49} & 37.56 & \textbf{38.38} & 37.48 & 37.17 & \underline{38.18} \\
\bottomrule
\end{tabular}
\end{table*}

\section{Experiments}
\subsection{Experimental Setup}
We base our experiments on NLLB-200 (No Language Left Behind) \cite{nllbteam2022languageleftbehindscaling}, a massively multilingual translation model that covers 200 languages. In particular, we use the distilled 600M and 1.3B parameter variants. NLLB is an encoder-decoder Transformer \cite{NIPS2017_3f5ee243} with language-specific tokens that specify the target language during generation. We use the NLLB-MD dataset \cite{ev, nllbteam2022languageleftbehindscaling} in all our experiments, and follow the provided split of parallel train, dev, and test sets with respective sizes of 6000, 1310, and 1600 sentences. We discard the target side during training., hence the self-supervised approach.
We implement our method in PyTorch \cite{paszke2019pytorchimperativestylehighperformance} using the Hugging Face Transformers library \cite{wolf-etal-2020-transformers}. We optimize with AdamW \cite{kingma2017adammethodstochasticoptimization} using a learning rate of $2 \times 10^{-6}$ for 2 epochs with batch size 2, group size $K{=}4$, KL coefficient $\beta{=}0.04$, clip parameter $\epsilon_c{=}0.2$, and a reference update frequency of 16 steps. During generation, we use nucleus sampling with temperature scaling (temperature 1.8, top-$k$ 100, top-$p$ 0.95) to encourage diverse candidate generation.
Our study focuses on the following languages: Central Aymara, Friulian, Wolof, Bhojpuri, Dyula and Russian. We use the GoldFish \cite{chang-etal-2024-goldfish} monolingual model family to evaluate the fluency of our system's forward translations for each language.

\subsection{Baselines}
\paragraph{Unsupervised Machine Translation
Using Monolingual Corpora Only (round-trip / cycle training).}
We include \citet{lample2018unsupervisedmachinetranslationusing} method with denoising and cycle-consistent reconstruction objectives, augmented with an adversarial discriminator. Models are trained for $2$ epochs with batch size $2$. We use separate learning rates for the encoder/decoder and discriminator: $\text{LR}_{\text{enc/dec}}=3\cdot10^{-5}$ and $\text{LR}_{\text{disc}}=5\cdot10^{-5}$. Objective weights are set to $\lambda_{\text{auto}}=1.0$ (denoising auto-encoding), $\lambda_{\text{cd}}=1.0$ (cycle/round-trip reconstruction), and $\lambda_{\text{adv}}=1.0$ (adversarial alignment). Input corruption uses word dropout $0.1$ and local shuffle with window size $k=3$.

\paragraph{Back-Translation} We implement the standard back-translation method of \citet{sennrich2016improvingneuralmachinetranslation}. 
Given monolingual target-language sentences $y$, we first generate synthetic English source sentences 
$\hat{x}$ using the pretrained NLLB model in the target$\rightarrow$English direction. 
The resulting synthetic parallel pairs $(\hat{x}, y)$ are then used to train the English$\rightarrow$target 
translation model using the standard cross-entropy objective. Synthetic translations are generated using the same pretrained NLLB model prior to reinforcement learning fine-tuning.

\paragraph{Round-Trip Translation with MLE}
Finally, a round-trip translation baseline \citep{10.1007/11941439_149} where sentences are translated from English to the target language and then translated back to the English using the same model parameters. The model is trained to reconstruct the original sentence after the two-step translation process using standard cross entropy objective.
\section{Results}

\subsection{Forward Translations Evaluation (English->Target)}

\paragraph{Translation Quality with chrF++} Following the original NLLB work \cite{nllbteam2022languageleftbehindscaling} in terms of low-resource language translation evaluation, Table~\ref{tab:main_results_roundtrip_rl} summarizes chrF++ scores on the forward translations before and after training. Across all languages and both model sizes, round-trip reinforcement training improves translation quality and fluency, yielding consistent absolute gains. We generally avoid model-based metrics like COMET \cite{rei2020cometneuralframeworkmt} or BERTScore \cite{zhang2020bertscoreevaluatingtextgeneration} for forward translations due to the lack of reliable scoring models for many of the languages we evaluate. However, for Russian—where a reliable COMET model is available—we additionally report COMET scores, which change from 0.84 before training to 0.86 after training.

Our method is particularly effective for very low-resource languages. With the 600M model, it achieves the best scores on Central Aymara (28.13), Wolof (23.96), and Dyula (22.50), outperforming BT by 0.79, 0.39, and 2.37 points respectively. It also consistently exceeds the round-trip (RT) baseline on these languages. With the 1.3B model, our method remains best on Central Aymara (28.73) and Dyula (22.97), while remaining competitive on Wolof (26.66), where UMT attains the top score (27.72).
For higher-resource languages (Russian, Friulian, Bhojpuri), BT is typically the strongest baseline, but the gap to our method remains small. Importantly, BT relies on target-to-English translation during training (tgt→English), effectively assuming access to a reverse-direction model or parallel signal. In contrast, our method only assumes the availability of an English monolingual corpus and does not require target→English supervision. Despite this weaker assumption, our approach remains competitive with BT for very low-resource languages like Dyula, Central Aymara and Wolof. Notably, our method consistently improves over the vanilla round-trip (RT) baseline across all languages and model sizes.
On average, our method achieves the best performance with the 600M model (36.49), outperforming BT by 0.31 points. With the 1.3B model, it remains competitive (38.18), within 0.20 of the strongest method (BT at 38.38). Finally, UMT exhibits less consistent behavior across languages and model sizes, which is plausibly explained by its more complex optimization objective that combines auto-encoding, cross-domain, and adversarial losses.

\begin{table}[t]
\centering
\small
\setlength{\tabcolsep}{12pt}
\renewcommand{\arraystretch}{1}
\begin{tabular}{lrrr}
\hline
Language & Before & After & \boldmath$\Delta$ \\
\hline
Central Aymara & -20.03 & -19.87 & \pos{0.16} \\
Friulian       & -19.23 & -19.09 & \pos{0.14} \\
Russian        & -15.64 & -15.37 & \pos{0.27} \\
Wolof          & -18.26 & -18.07 & \pos{0.19} \\
Bhojpuri       & -14.93 & -14.97 & \neg{-0.04} \\
Dyula          & -17.53 & -16.43 & \pos{1.09} \\
\hline
\end{tabular}
\caption{Goldfish model log-probability scores (natural log) before (vanilla) and after training, with improvement $\Delta = \text{trained} - \text{vanilla}$.}
\label{tab:goldfish_logprob_scores}
\end{table}

\paragraph{Translation Fluency} Since we only reward the round-trip translation, it might be possible that the model generates nonsensical forward translation text, despite having an improved translation quality according to chrF++. To measure this, we measure target-side fluency using a monolingual LM under the \emph{Goldfish} \citet{chang-etal-2024-goldfish} language model, reporting average log-probability (natural log) before and after training (Table~\ref{tab:goldfish_logprob_scores}). Across five of the six languages, training improves LM likelihood (i.e., makes it less negative), consistent with more fluent and model-preferred output distributions, confirming that our method does not degrade fluency. For Bhojpuri, we think that due to the higher morphological variability in this language, the system does not improve fluency.

These scores suggest that the round-trip reinforcement objective enhances not only adequacy, as captured by chrF++, but also surface-level well-formedness and overall likelihood according to an external scorer. These results are further supported by the qualitative analysis of the translation output discussed in a later section.

\subsection{Validation Curves For Forward Translations (English->Target)}

\begin{figure}[t]
    \centering
    \begin{subcaptionblock}{0.49\linewidth}
        \centering
        \includegraphics[width=\linewidth]{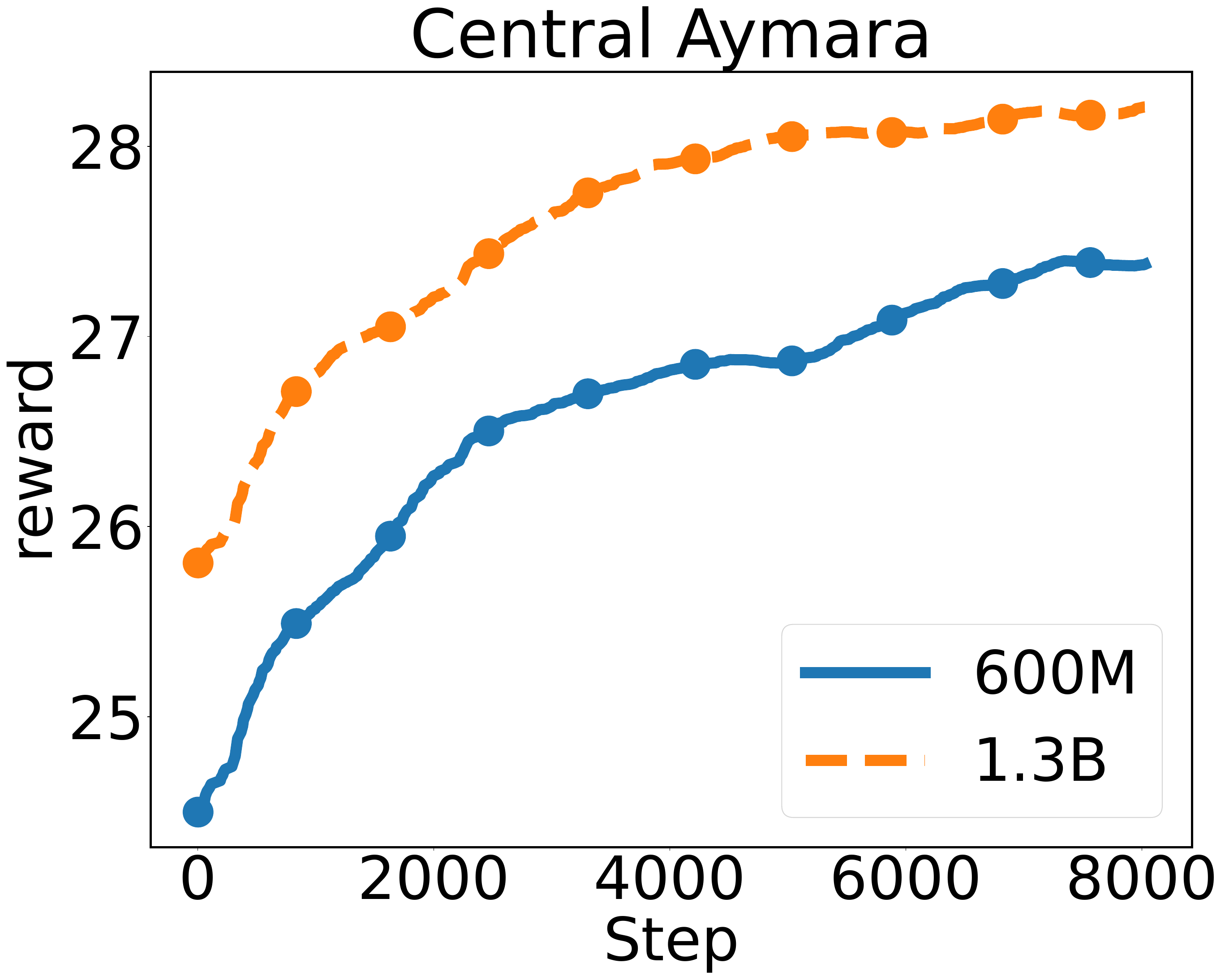}
    \end{subcaptionblock}
    \hfill
    \begin{subcaptionblock}{0.49\linewidth}
        \centering
        \includegraphics[width=\linewidth]{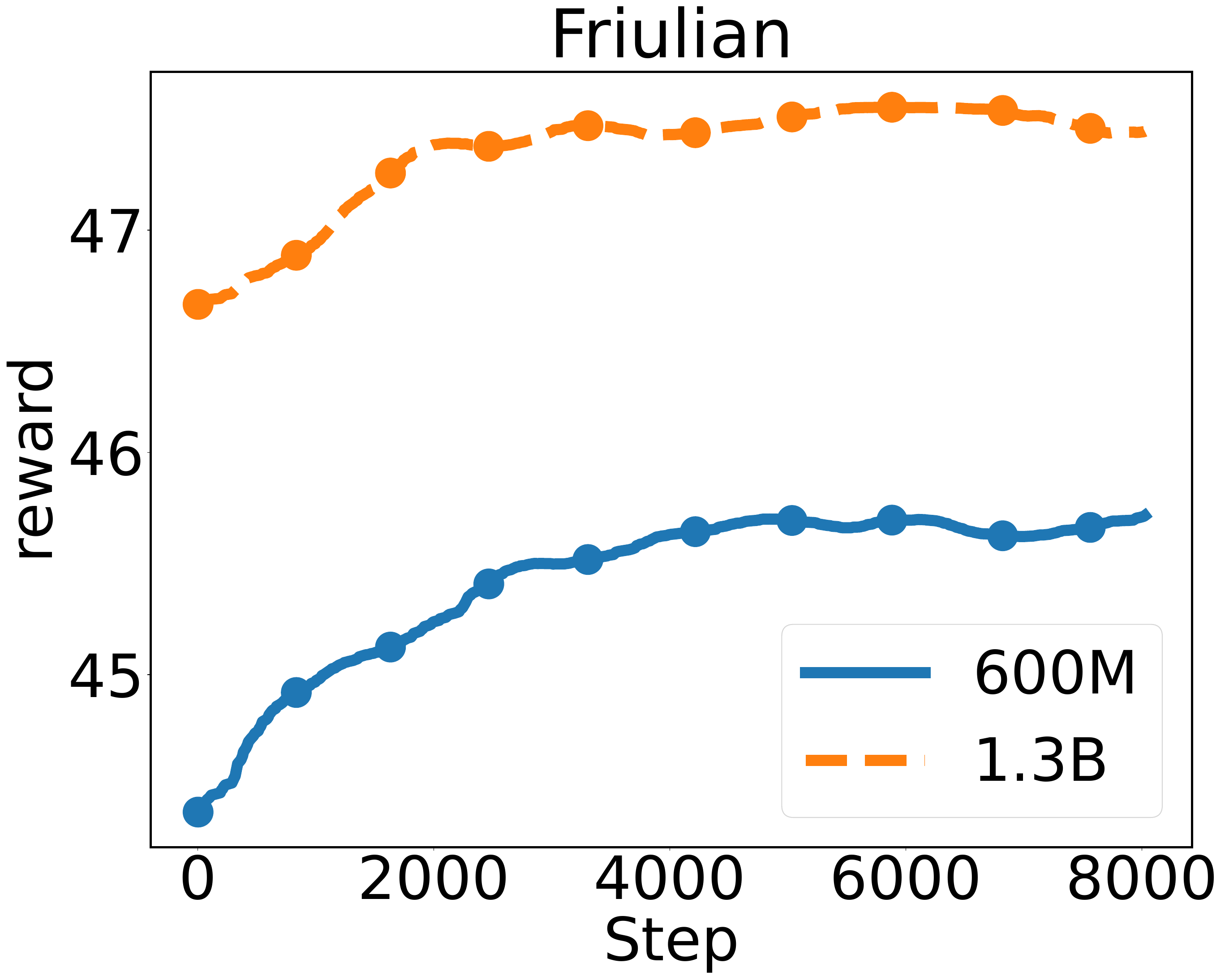}
    \end{subcaptionblock}

    \vspace{0.7em}

    \begin{subcaptionblock}{0.49\linewidth}
        \centering
        \includegraphics[width=\linewidth]{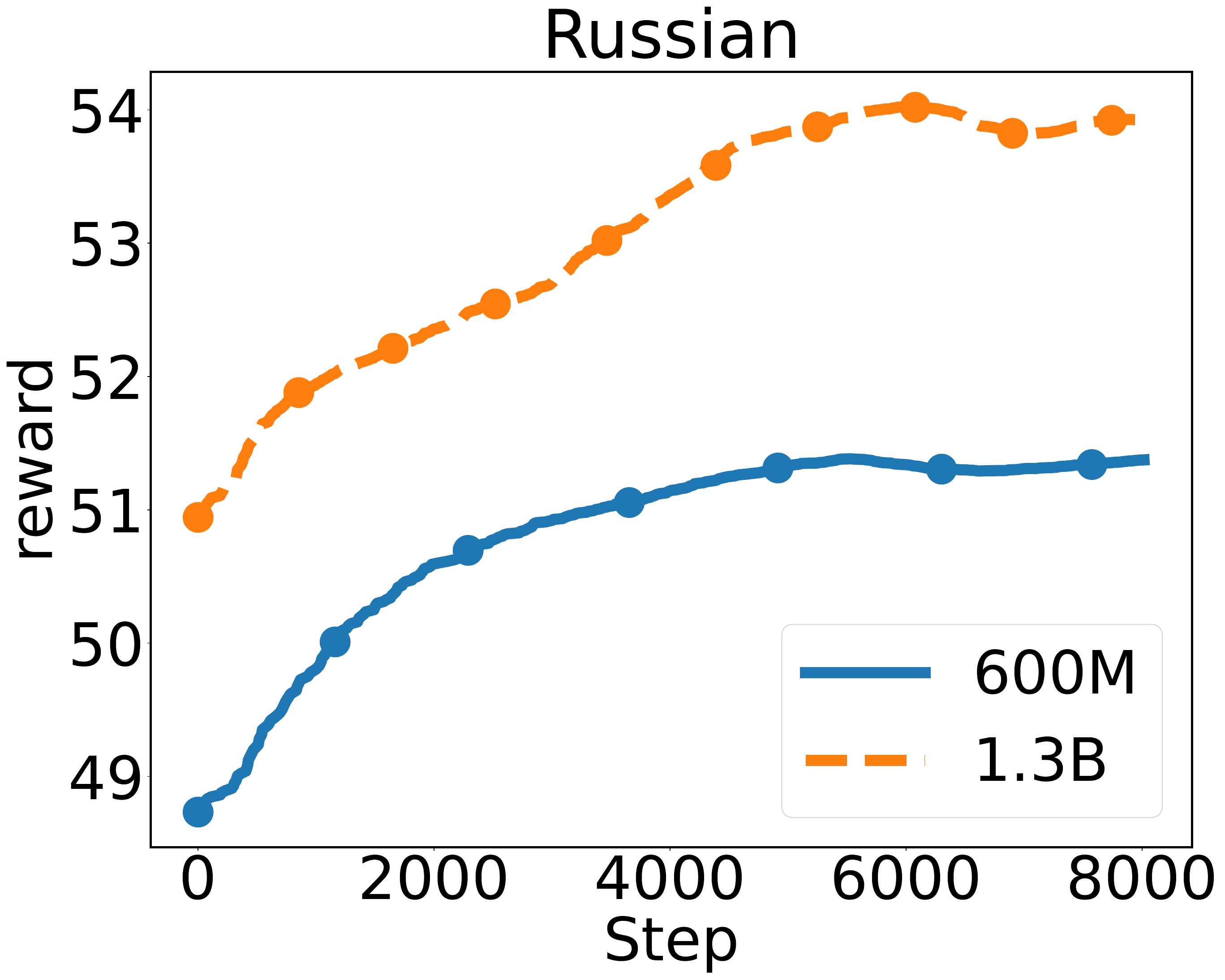}
    \end{subcaptionblock}
    \hfill
    \begin{subcaptionblock}{0.49\linewidth}
        \centering
        \includegraphics[width=\linewidth]{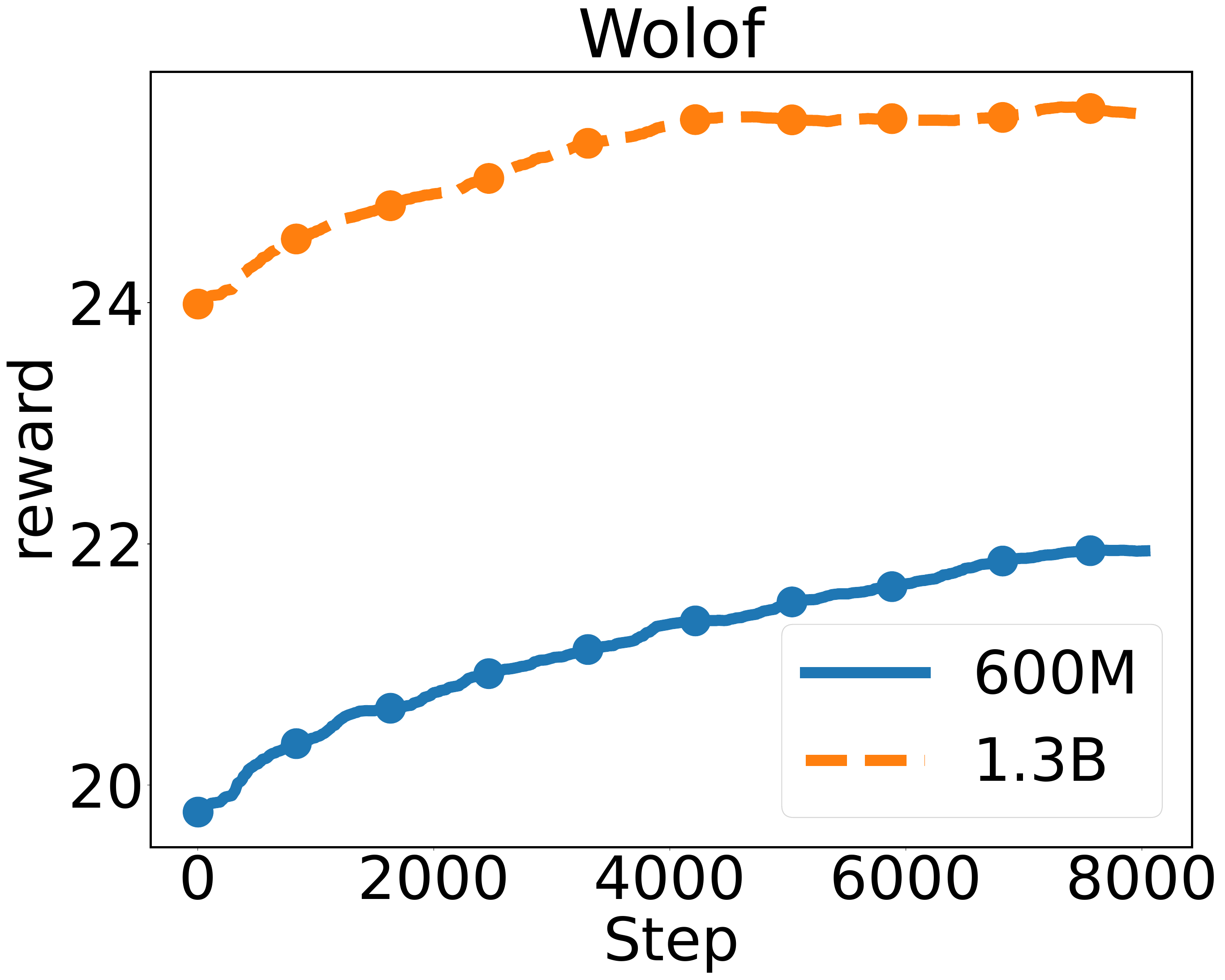}
    \end{subcaptionblock}
    \begin{subcaptionblock}{0.49\linewidth}
        \centering
        \includegraphics[width=\linewidth]{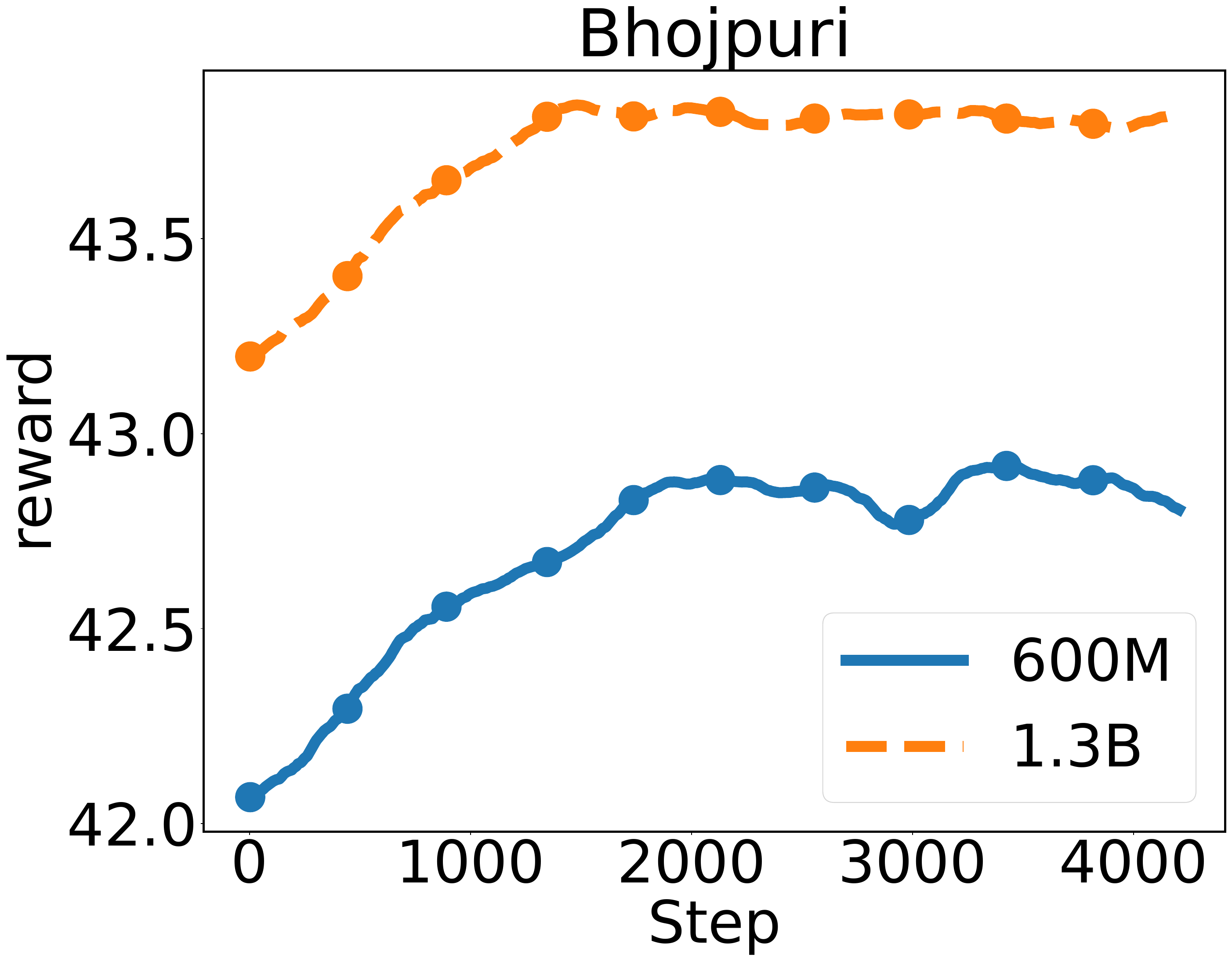}
    \end{subcaptionblock}
    \begin{subcaptionblock}{0.49\linewidth}
        \centering
        \includegraphics[width=\linewidth]{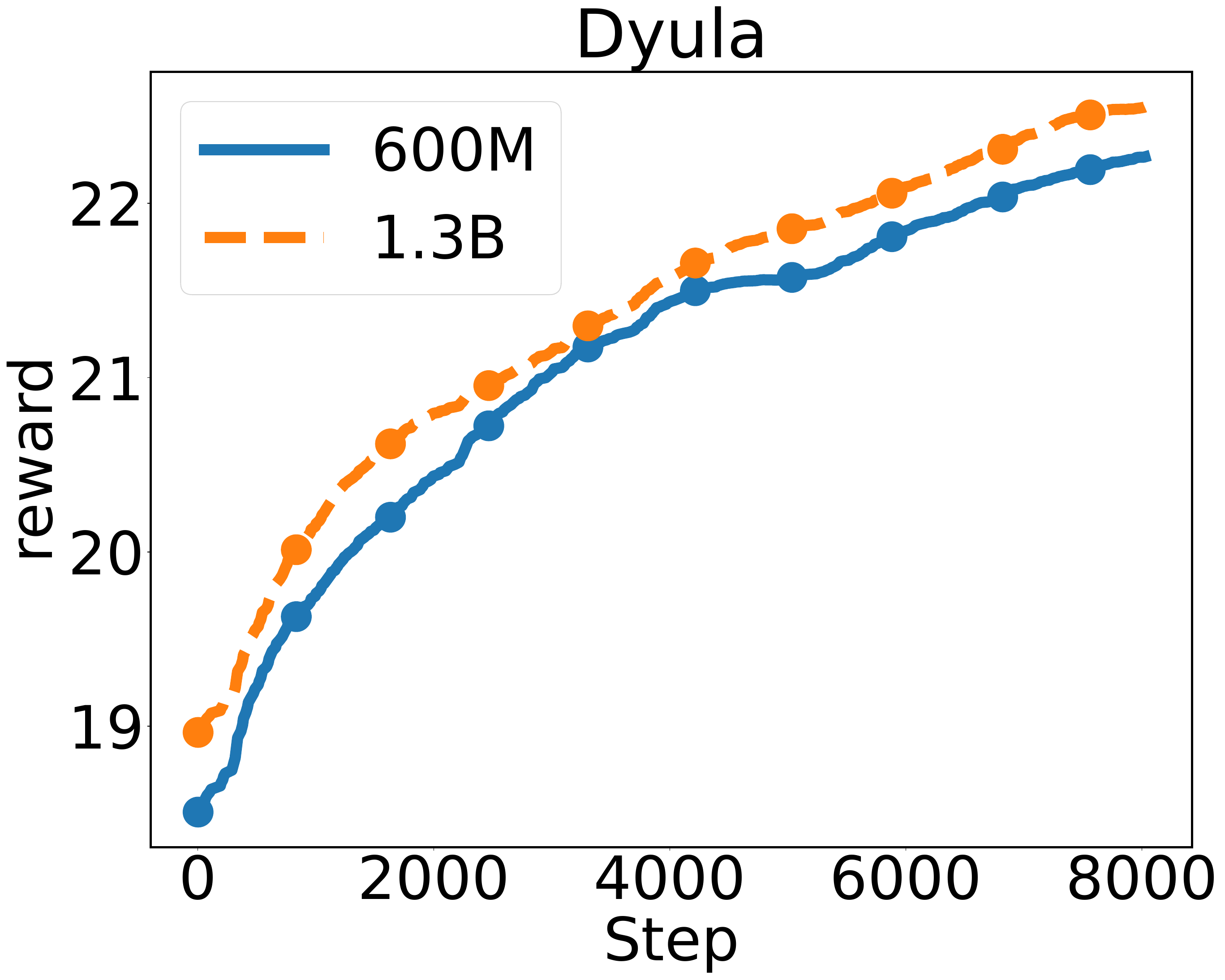}
    \end{subcaptionblock}
    \caption{Validation curves showing forward translation chrF++ scores (reward) for the six languages after 8K optimization steps.}
    \label{fig:reward-six-langs}
\end{figure}

Figure \ref{fig:reward-six-langs} shows the chrF++ scores curves on the out-of-distribution evaluation set of NLLB-MD for the six languages and both the 1.3B and 600M NLLB models. 
In all six languages, both models improve steadily over the course of training. The curves are smooth and approximately monotonic, without large oscillations, which indicates that the optimization procedure is stable.
Across all languages and throughout training, the 1.3B model consistently outperforms the 600M model. The 1.3B curves start from higher chrF++ scores and maintain a similar margin over the 600M curves at every step. The gap remains roughly stable over time, which implies that scaling model size gives better improvements. Also, The overall shape of the curves is similar across languages, but the absolute gains differ.

\subsection{Reward Function Ablation Study}\label{sec:reward-ablation}

We next study how the choice of intrinsic reward impacts training results. We compare three reward functions: (i) a uniform weighted combination of chrF++ and BLEU (``BLEU+chrF++''), (ii) BLEU only, and (iii) chrF++ only. In all cases, we keep the training procedure and hyperparameters fixed and report the \emph{forward translation} chrF++ score before and after adaptation on each language and model size.

\begin{figure}[t]
  \centering
  \includegraphics[width=\columnwidth]{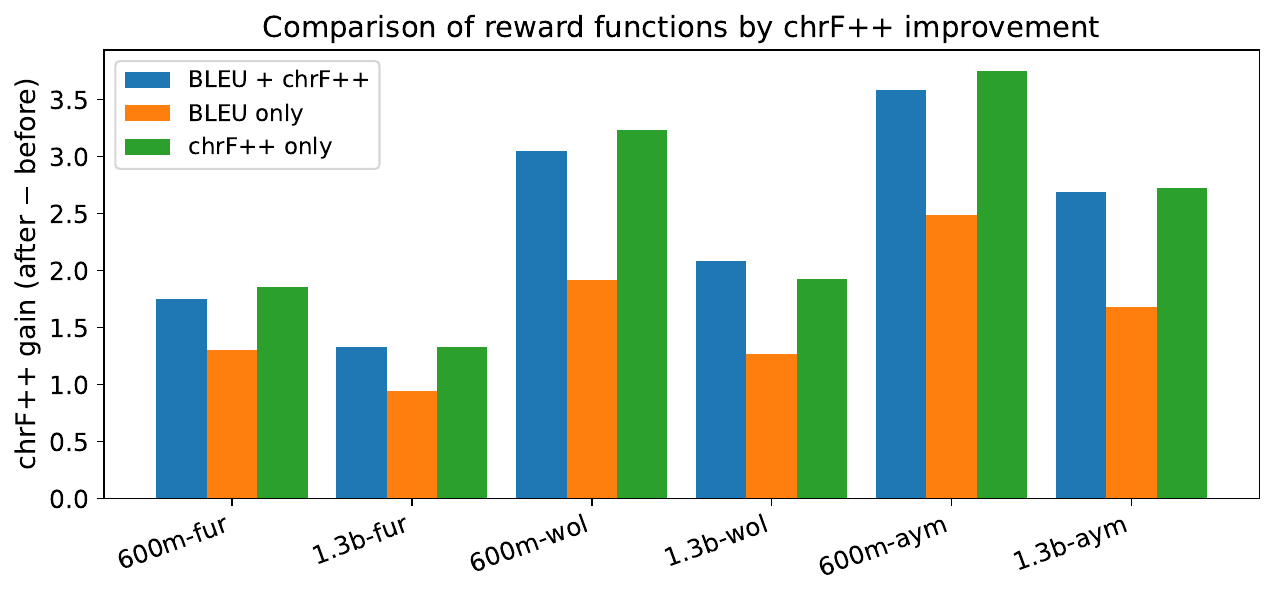}
  \caption{forward translations chrF++ gain (after -- before) for different reward functions across model/language settings. (fur=Friulian, wol=Wolof, aym=Central Aymara and rus=Russian)}
  \label{fig:chrf-gain}
\end{figure}

Figure~\ref{fig:chrf-gain} summarizes the chrF++ \emph{gain} (after minus before). Across all six settings, BLEU-only rewards consistently improve performance but yield substantially smaller gains than rewards that include chrF++. Averaged over all conditions, BLEU-only increases chrF++ by +1.60, whereas BLEU+chrF++ yields +2.41 and chrF++-only yields +2.47.

\begin{figure}[t]
    \centering
    \begin{subcaptionblock}{0.49\linewidth}
        \centering
        \includegraphics[width=\linewidth]{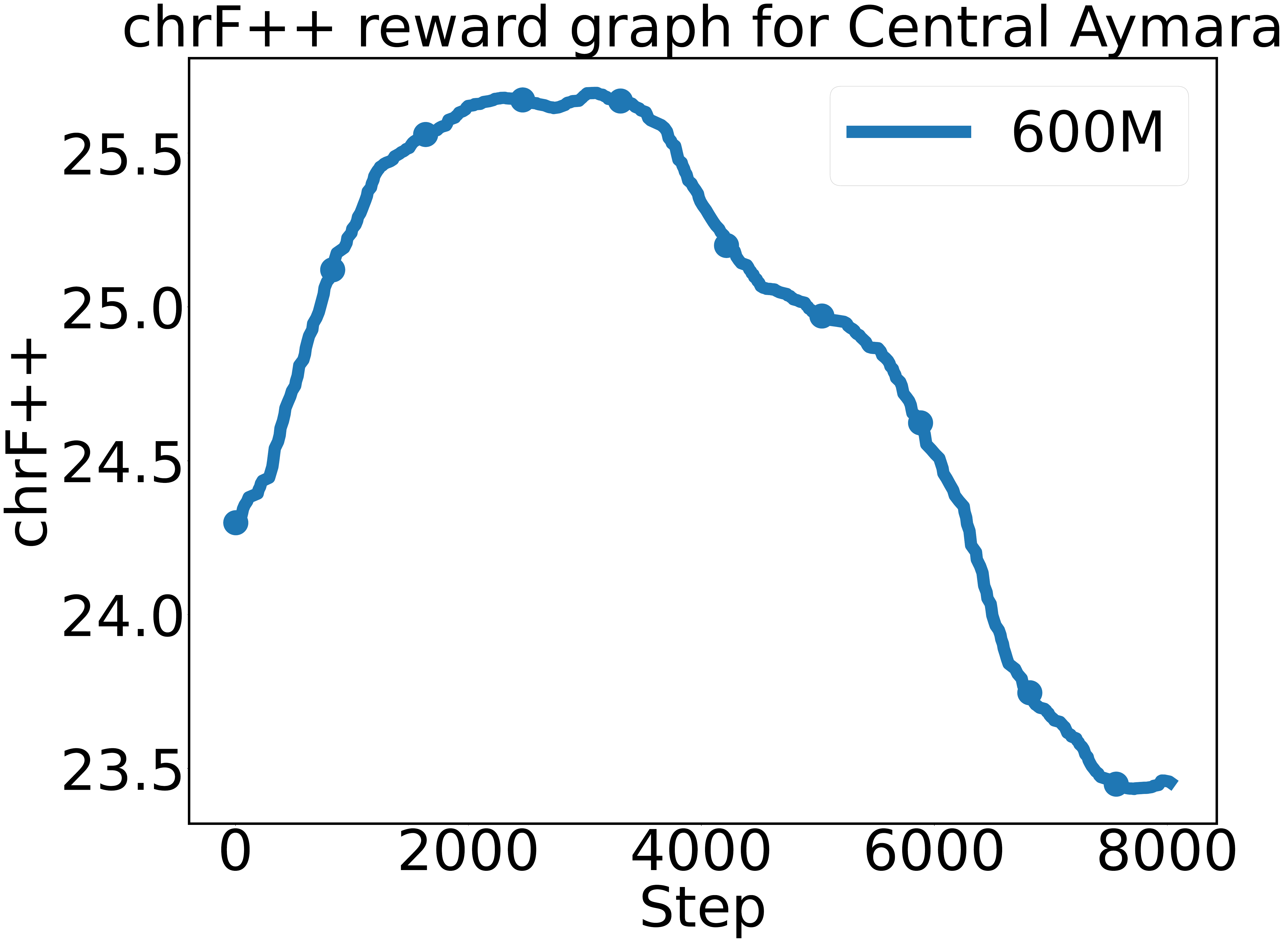}
    \end{subcaptionblock}
    \hfill
    \begin{subcaptionblock}{0.49\linewidth}
        \centering
        \includegraphics[width=\linewidth]{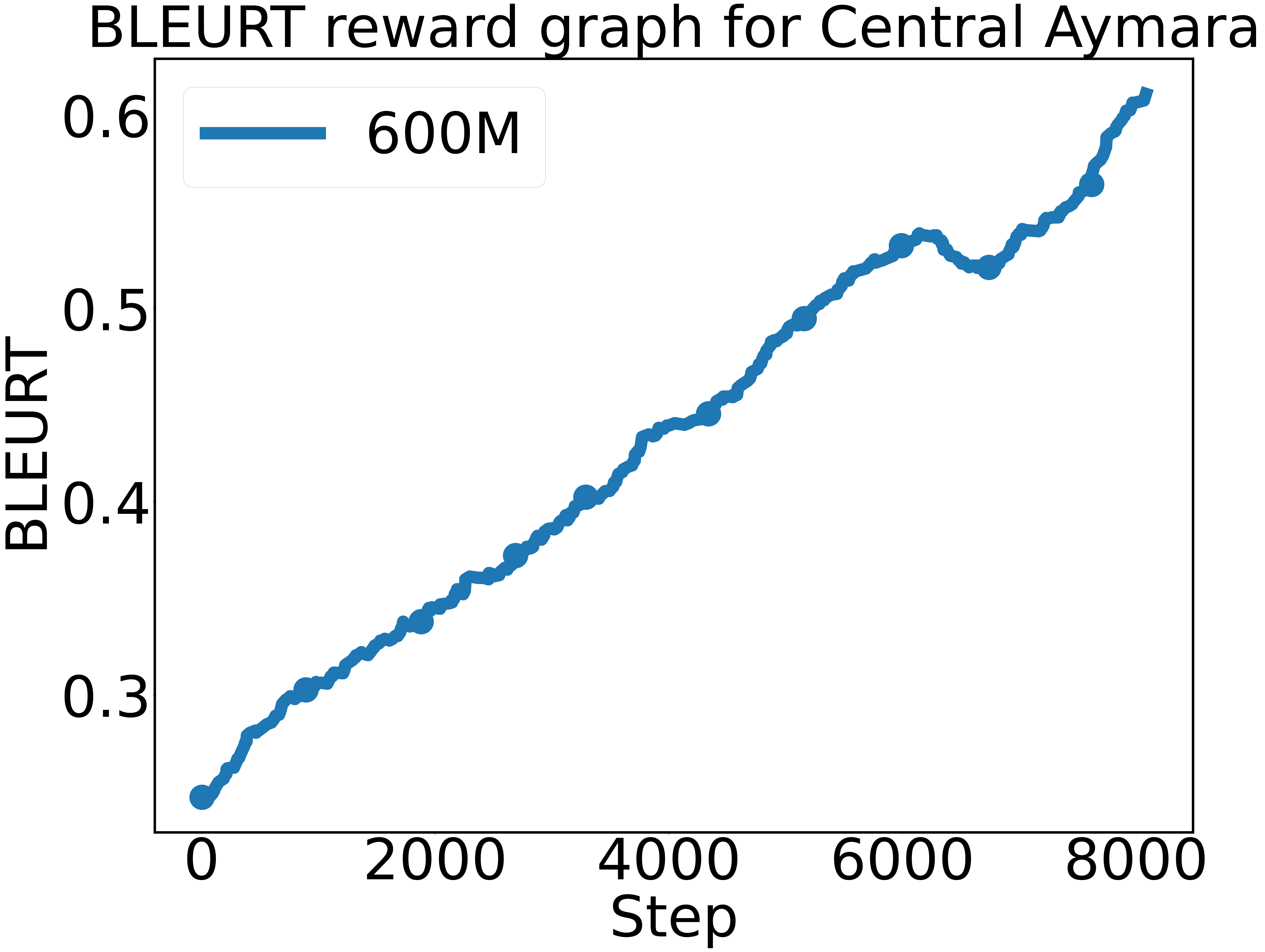}
    \end{subcaptionblock}

    \caption{Results after training the 600M model with BLEURT as the main back-translation reward on Central Aymara. Left: chrF++ reward graph on English->Target direction on the validation set. Right: BLEURT Scores during training between the source and generated English back-translation.}
    \label{fig:bleurt}
\end{figure}

This trend is especially pronounced for the most challenging pairs (e.g., Wolof and Central Aymara), where BLEU-only rewards provide a weaker learning signal under sparse n-gram matching. By contrast, chrF++ provides a smoother character-level similarity objective that appears better aligned with low-resource and morphologically rich settings, leading to the strongest and most consistent improvements (best in 4/6 conditions). Adding BLEU to chrF++ is competitive and occasionally beneficial (notably for \texttt{1.3b-wol}), but does not uniformly outperform chrF++ alone, suggesting that word-level precision signals can sometimes be redundant or slightly misaligned with the round-trip consistency objective. Finally, as shown in figure \ref{fig:bleurt}, we experimented with BLEURT \citep{sellam2020bleurtlearningrobustmetrics} as the reward in the English reconstruction step, comparing reconstructed sentences to the original English source. However, training became unstable: the model rapidly exploited the learned metric, producing outputs that achieved high BLEURT scores while true translation quality deteriorated, leading to a collapse in forward translations chrF++ scores on the validation set. This behavior is consistent with reward over-optimization \citep{paulus2017deepreinforcedmodelabstractive, lazaridou-etal-2020-multi} when using learned metrics as training objectives.

\begin{figure}[t]
    \centering
    \begin{subcaptionblock}{0.49\linewidth}
        \centering
        \includegraphics[width=\linewidth]{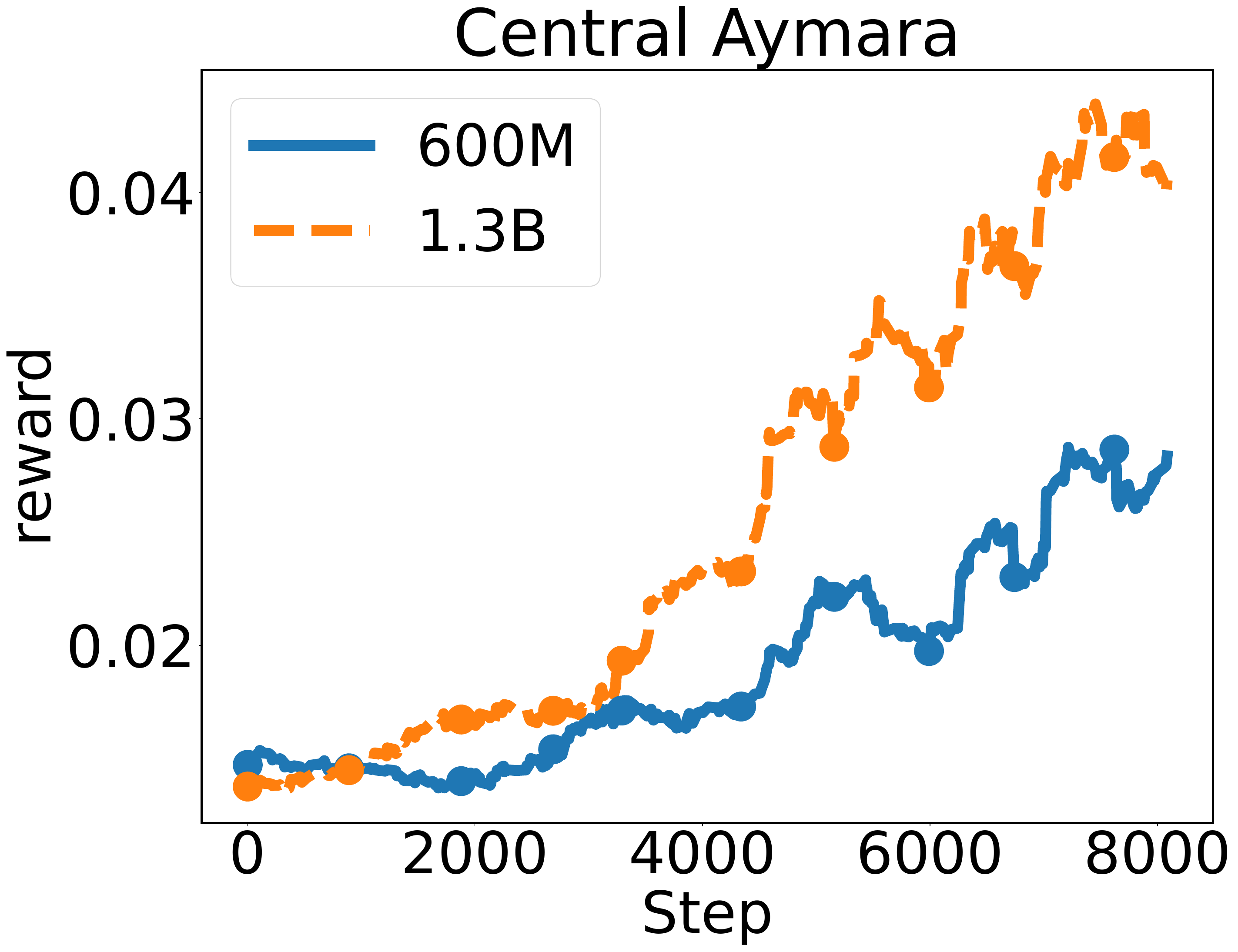}
    \end{subcaptionblock}
    \hfill
    \begin{subcaptionblock}{0.49\linewidth}
        \centering
        \includegraphics[width=\linewidth]{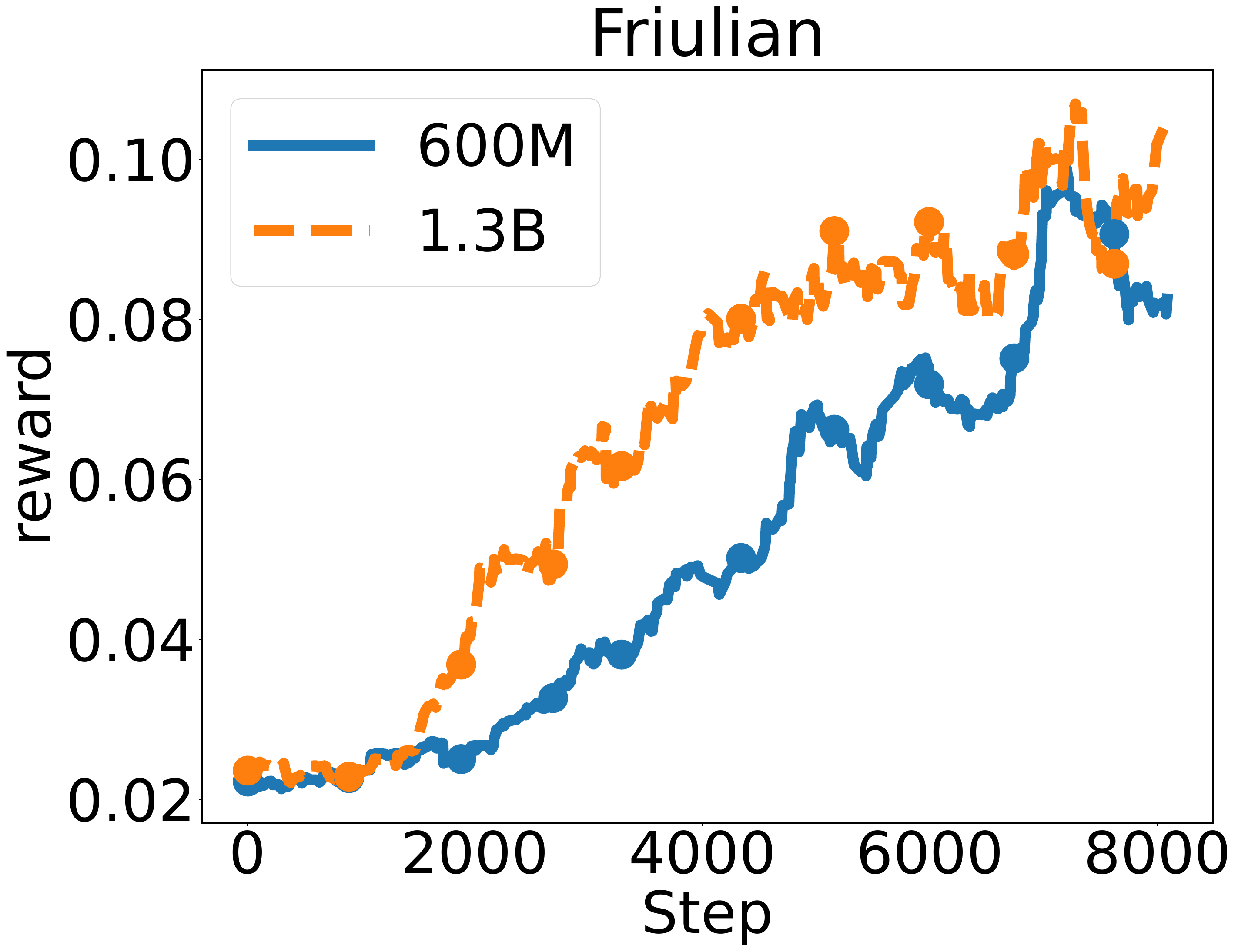}
    \end{subcaptionblock}

    \vspace{0.7em}

    \begin{subcaptionblock}{0.49\linewidth}
        \centering
        \includegraphics[width=\linewidth]{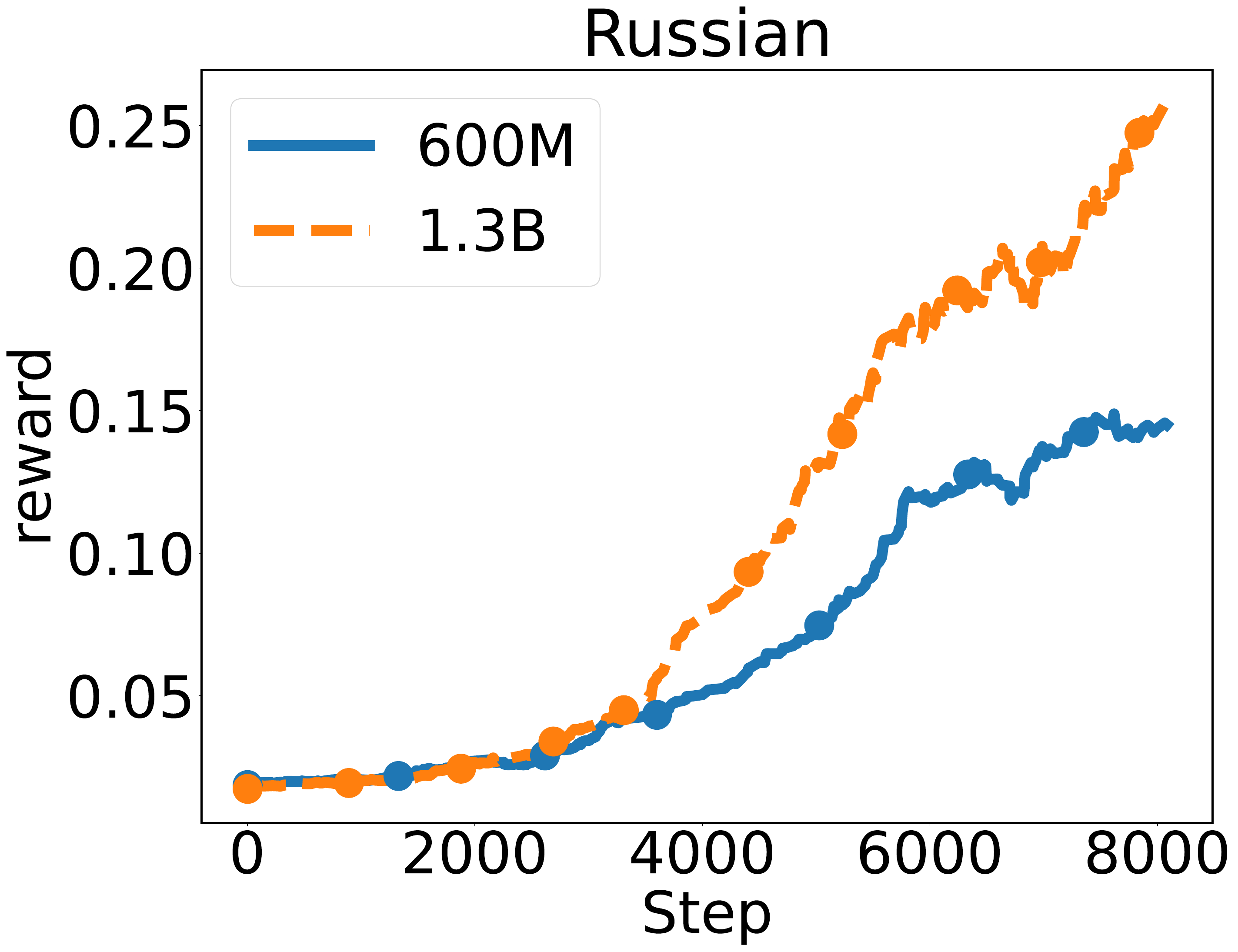}
    \end{subcaptionblock}
    \hfill
    \begin{subcaptionblock}{0.49\linewidth}
        \centering
        \includegraphics[width=\linewidth]{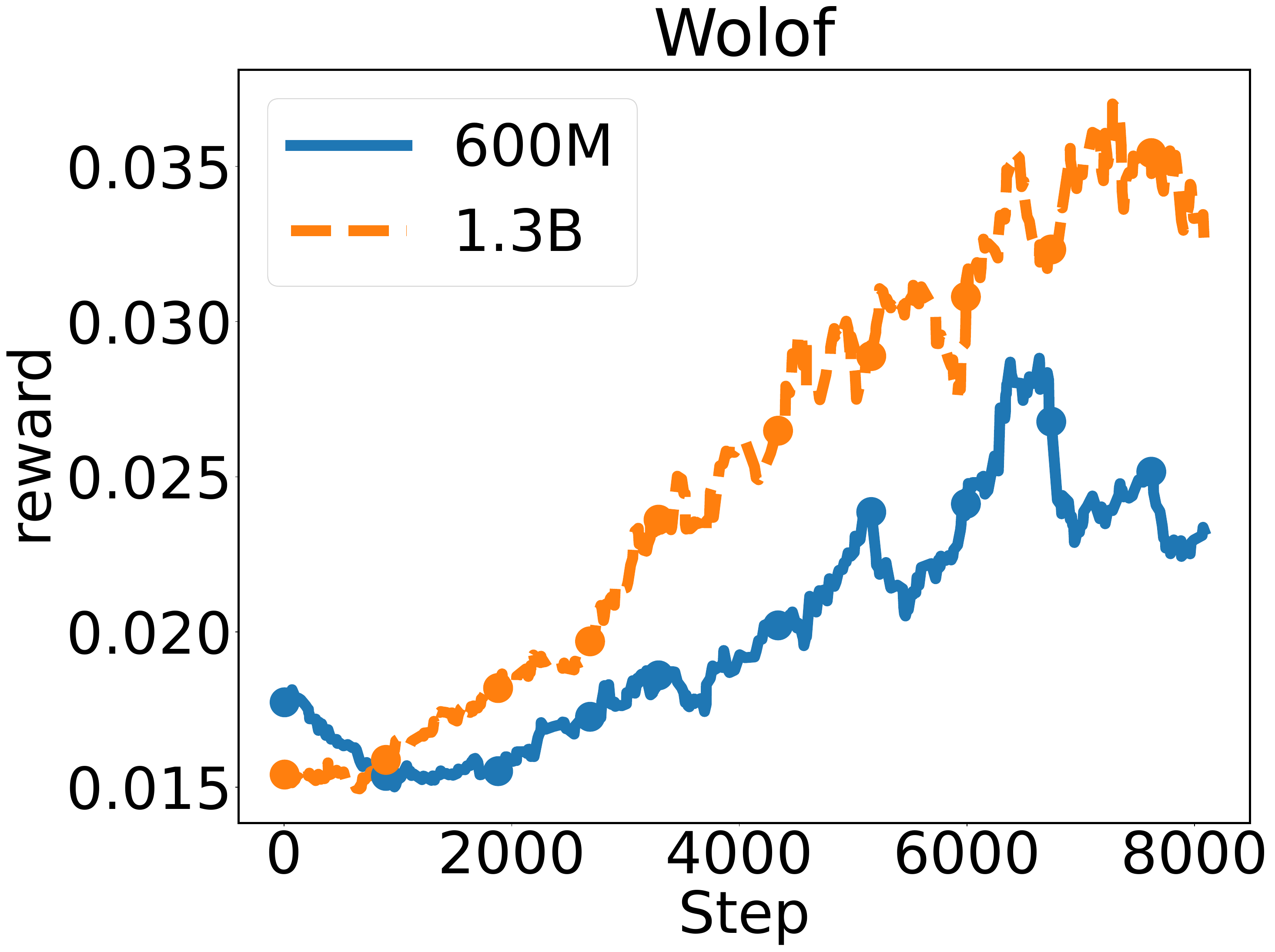}
    \end{subcaptionblock}
    \begin{subcaptionblock}{0.49\linewidth}
        \centering
        \includegraphics[width=\linewidth]{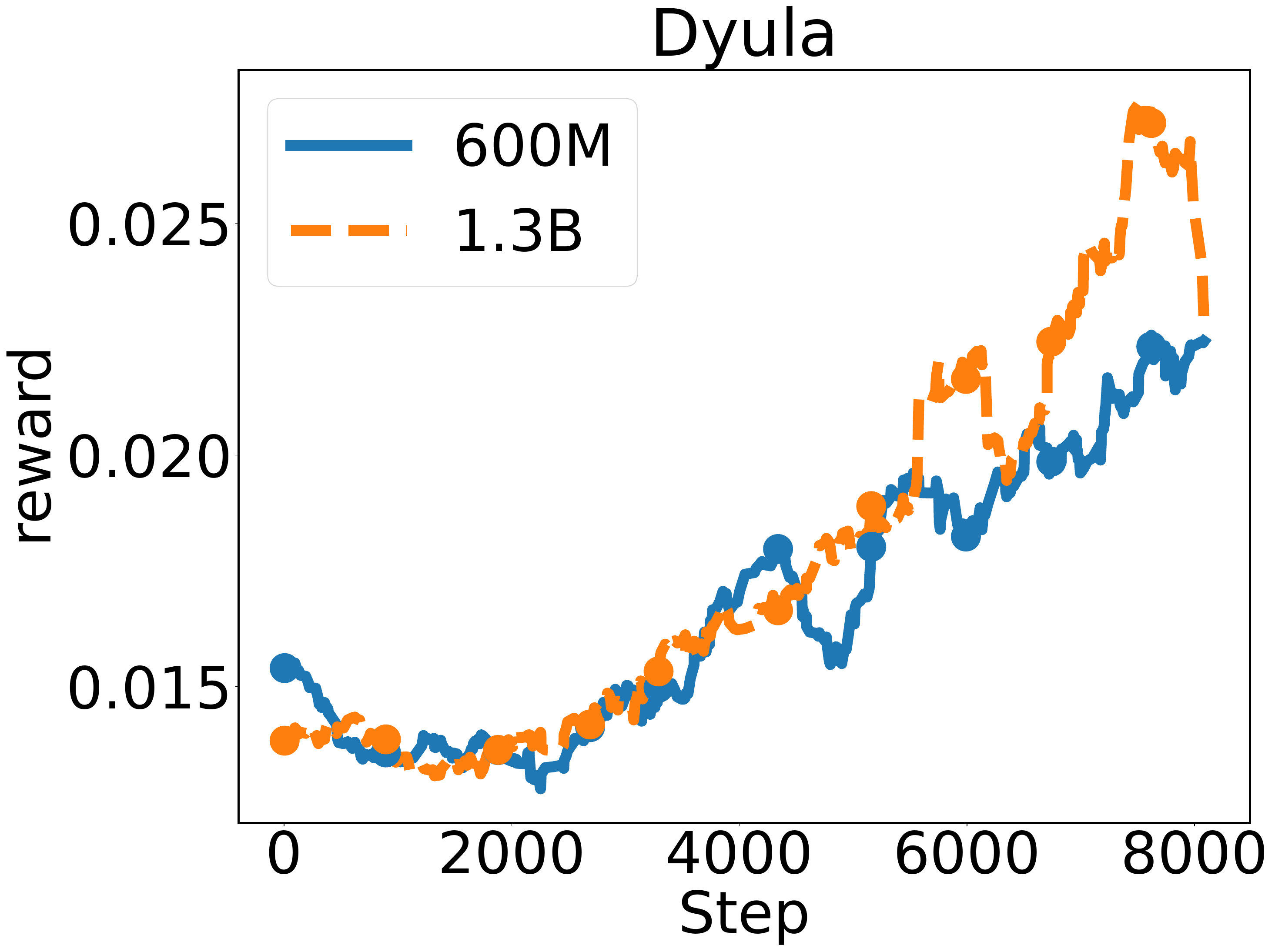}
    \end{subcaptionblock}
    \begin{subcaptionblock}{0.49\linewidth}
        \centering
        \includegraphics[width=\linewidth]{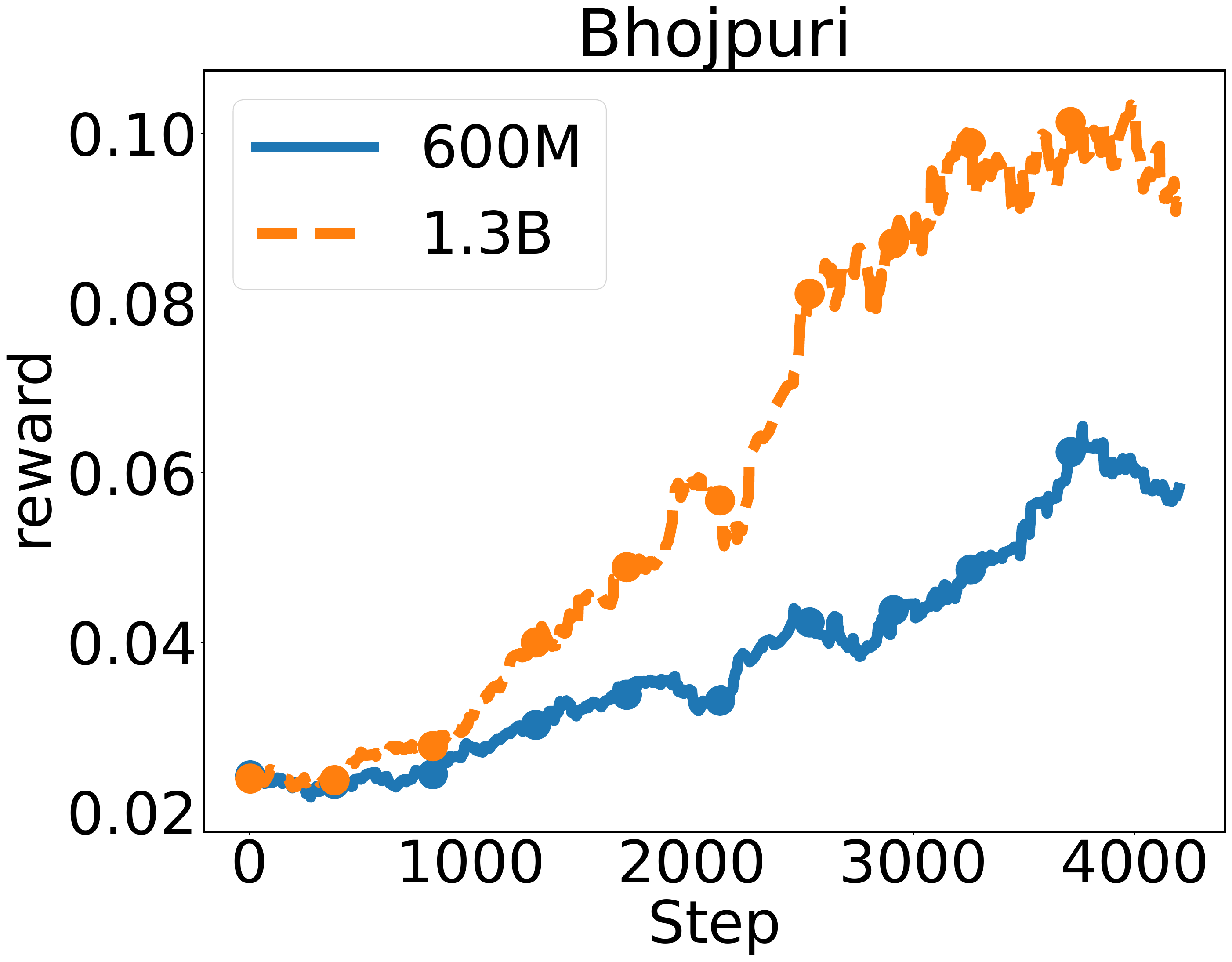}
    \end{subcaptionblock}
    \caption{Training curves showing the tgt->English backward translation BLEU scores (reward) during training for the 1.3B and 600M parameter models for the 6 Languages across 8K optimization steps.}
    \label{fig:reward-backtranslation}
\end{figure}

\begin{table}[t]
\centering
\scriptsize
\setlength{\tabcolsep}{5.5pt} 
\renewcommand{\arraystretch}{1} 
\begin{tabular}{lccccccc}
\toprule
\textbf{Target} &
\multicolumn{3}{c}{\textbf{Before}} &
\multicolumn{3}{c}{\textbf{After}} &
\textbf{$\Delta$F1} \\
\cmidrule(lr){2-4}\cmidrule(lr){5-7}
 & P & R & F1 & P & R & F1 & \\
\midrule
Central Aymara & 0.48 & 0.49 & 0.48 & 0.54 & 0.57 & 0.55 & \pos{+0.07} \\
Friulian & 0.78 & 0.73 & 0.75 & 0.78 & 0.76 & 0.77 & \pos{+0.02} \\
Russian & 0.77 & 0.69 & 0.73 & 0.80 & 0.77 & 0.79 & \pos{+0.06} \\
Wolof & 0.53 & 0.48 & 0.51 & 0.54 & 0.55 & 0.55 & \pos{+0.04} \\
Dyula & 0.47 & 0.35 & 0.40 & 0.56 & 0.56 & 0.56 & \pos{+0.16} \\
Bhojpuri & 0.80 & 0.77 & 0.79 & 0.77 & 0.77 & 0.77 & \neg{-0.02} \\
\bottomrule
\end{tabular}
\caption{Mean Precision/Recall/F1 for vanilla vs.\ trained models English Back-Translations using BERTScore.}
\label{tab:bertscore_results}
\end{table}

\begin{table*}[ht!]
\centering
\scriptsize
\begin{tabularx}{\textwidth}{@{}l l c X@{}}
\toprule
\textbf{Output Source} & \textbf{Language} & \textbf{chrF++} & \textbf{Text} \\
\midrule

Source (English)
& English
& N/A
& That's so awesome! Which one is it? \\
\cmidrule(lr){4-4}

Reference
& Central Aymara
& N/A
& ¡Ukax wali musphkayawa! ¿Kawkïrisa? \\
\cmidrule(lr){4-4}

Base Model
& Central Aymara
& 35.32
& \mismatch{¡Wali} \mismatch{sumäskiwa}! \improved{¿Kawkïris} \mismatch{uka?} \\
\cmidrule(lr){4-4}

Trained Model
& Central Aymara
& 64.09
& \match{¡Ukax} \match{wali} \improved{musparkañawa}!
  \improved{¿Kawkïris} \mismatch{ukaxa?} \\
\midrule

Source (English)
& English
& N/A
& However, it was also of great importance "for a country from which we are separated by only two nations," she said. \\
\cmidrule(lr){4-4}

Reference
& Friulian
& N/A
& Dut câs, e jere ancje di grande impuartance "par une nazion de cuâl o sin separâts dome di dôs nazions," e à dite. \\
\cmidrule(lr){4-4}

Base Model
& Friulian
& 16.61
& \mismatch{"O vin di fâ alc par fâ cressi la nestre identitât",}
  \match{e à dite}
  \mismatch{la ministra.} \\
\cmidrule(lr){4-4}

Trained Model
& Friulian
& 70.46
& \match{Dut câs, e jere ancje di grande impuartance}
  \improved{"par un paîs di dulà che o sin separâts dome di dôs nazions",}
  \match{e à dit} \mismatch{jê.} \\
\midrule

Source (English)
& English
& N/A
& Rheumatic illnesses in particular rheumatic arthritis and ankylosing spondylitis (so called Bechterew's disease) \\
\cmidrule(lr){4-4}

Reference
& Wolof
& N/A
& Jàngoroy Rheumatik rawatina arthritis rheumatik ak ankylosing spondytilis (nu duppe ko jàngoro Bechterew) \\
\cmidrule(lr){4-4}

Base Model
& Wolof
& 5.84
& \mismatch{Jàmm yuy néew-ji-doole yuy Jàmm yuy néew-ji-doole yuy}  \\
\cmidrule(lr){4-4}

Trained Model
& Wolof
& 42.22
& \mismatch{Yàmm yu }
  \match{réwmatik, rawatina aartritis réwmatik ak spondylitis ankylosing}
  \mismatch{(yite bi ñuy wax jàngoroy} 
  \match{Bechterew)} \\
\midrule

Source (English)
& English
& N/A
& Sounds great to me, I'm so excited! What is the horse's name? \\
\cmidrule(lr){4-4}

Reference
& Russian
& N/A
& Zvuchit zdorovo, ya tak rad! Kak zovut konya? \\
\cmidrule(lr){4-4}

Base Model
& Russian
& 21.32
& \mismatch{-} \match{Kak} \match{zovut} \mismatch{loshadey?} \\
\cmidrule(lr){4-4}

Trained Model
& Russian
& 74.52
& \match{Zvuchit} \match{zdorovo,} \match{ya} \match{tak} \improved{rada!}
  \match{Kak} \match{zovut} \mismatch{loshadey?} \\
\bottomrule
\end{tabularx}
\caption{Example outputs for Source (English), Reference translations, Base Model, and Trained model across four languages.
Tokens in green match the reference, tokens in red differ from the reference, and tokens in blue mark segments where the trained model is closer to the reference than the base model. chrF++ scores are shown for model outputs with respect to the Reference sentence.}
\label{tab:qualitative}
\end{table*}

\subsection{Back-Translation Evaluation (Target->English)}
Figure~\ref{fig:reward-backtranslation} shows backward-translation rewards (target~$\rightarrow$~English) over training steps for the 600M and 1.3B models. Both models exhibit a clear upward trend across all six languages, indicating that continued training consistently improves backward translation quality rather than leading to early overfitting. The curves are noisier than in the forward setting, which is expected given the higher variance of the backward reward signal, but the overall trajectory is monotonic. It is clear that scaling model capacity yields consistent benefits for backward translation, and that the adaptation procedure is able to exploit the additional capacity rather than saturating at the smaller model’s performance level. Finally, this aligns with the forward translation reward graphs and our hypothesis is sound: RL optimization for the roundtrip back-translation chrF++ scores improves translation in both directions.

\paragraph{BERTScore}
To validate our findings beyond chrF++ and BLEU, which are used as reward signals during training, we also report results using BERTScore \cite{zhang2020bertscoreevaluatingtextgeneration}. Table \ref{tab:bertscore_results} shows the BERTScores computed on the back-translated English sentences before and after training. The trained model improves BERTScore F1 in five of the six cases, with gains ranging from +0.02 (Friulian) to +0.16 (Dyula). The largest improvements occur for Dyula (+0.16). The decrease in Bhojpuri again (-0.02) can be due to weaker bilingual alignment and a higher morphological variability. However, the decrease is very small. The substantial gain for Dyula reflects improvements in both precision and recall, suggesting that training yields back-translations that are both more semantically complete and better aligned with the reference under this embedding-based metric.

\subsection{Qualitative Analysis On Translation Outputs}

Table~\ref{tab:qualitative} presents a qualitative comparison between the 1.3B base model and the trained model in four evaluation languages (one example each; additional samples appear in Appendix~\ref{app:qualitative}). Overall, the trained model produces outputs that are substantially closer to the human reference.
The examples indicate that self-supervised training improves both (i) content coverage and faithfulness (notably in Friulian and Wolof) and (ii) surface realization, including idiomatic openers and morphology (Central Aymara and Russian). The remaining errors primarily involve fine-grained lexical choice and function-word control. Despite relying only on monolingual reward signals, the model improves translation quality in low-resource languages without direct supervision.

\section{Conclusion}
We propose a self-supervised RL fine-tuning approach for low-resource machine translation using round-trip bootstrapping with NLLB. The model translates English into a target language and is rewarded based on the quality of the reconstructed English using an intrinsic metric like chrF++. GRPO training yields consistent chrF++ improvements on an out-of-distribution NLLB-MD test set for both 600M and 1.3B models. Qualitatively, the trained models reduce common low-resource errors such as repetition and semantic drift, producing outputs that are more fluent and faithful. Ablations show chrF++ provides the most reliable learning signal, while combining BLEU is competitive but not consistently better. These results suggest pretrained multilingual models can self-improve very low-resource translation using only monolingual reward signals without parallel supervision.

\section{Limitations}
Our study has several limitations that point to clear directions for future work.

\paragraph{Scaling to larger NLLB checkpoints.}
Due to resource constraints, we only evaluate the distilled 600M and 1.3B NLLB variants, and do not test whether the same round-trip GRPO training remains stable and beneficial for substantially larger models (e.g., 3.3B NLLB).
Empirical scaling laws suggest that multilingual MT quality often improves predictably with model scale, and that scaling can change the effective capacity allocated to individual language pairs, including out-of-distribution behavior \citep{pmlr-v202-fernandes23a,kaplan2020scalinglawsneurallanguage}.

\paragraph{Broader low-resource language coverage within NLLB-MD.}
We focus on six languages and thus extending experiments to additional languages would strengthen conclusions about robustness across scripts, typology, and domain mismatch, and would help identify when round-trip rewards are most effective versus when they induce overly paraphrastic solutions.

\section{Ethical Considerations}
Our approach fine-tunes multilingual MT models using self-supervised, round-trip reinforcement learning (English$\rightarrow$target$\rightarrow$English) with automatic rewards (chrF++/BLEU). While this avoids collecting new human-labeled parallel data, it introduces several potential risks:

\paragraph{Amplification of model errors and biases.}
Because training relies on the model’s own generations, systematic errors (e.g., gender/role stereotypes, offensive terms, dialectal mismatches) can be reinforced. Biases present in pretraining or in monolingual English data may propagate into low-resource outputs.

\paragraph{Hallucinations and harmful content translation.}
RL may increase fluency while still hallucinating facts, altering named entities, or mistranslating safety-critical content (medical/legal). Additionally, improved MT quality could facilitate translation of toxic or hateful content into additional languages.

\paragraph{Mitigations.}
We mitigate these risks by (i) monitoring for degeneracy (repetition/copying) and reporting qualitative examples, (ii) evaluating with multiple metrics (chrF++, BLEU, BERTScore on back-translations, and target-side fluency), (iii) limiting training data to vetted corpora and documenting compute settings for transparency.

\bibliography{custom}

@inproceedings{he-etal-2024-recovery,
    title = "Recovery Should Never Deviate from Ground Truth: Mitigating Exposure Bias in Neural Machine Translation",
    author = "He, Jianfei  and
      Sun, Shichao  and
      Jia, Xiaohua  and
      Li, Wenjie",
    editor = "Scarton, Carolina  and
      Prescott, Charlotte  and
      Bayliss, Chris  and
      Oakley, Chris  and
      Wright, Joanna  and
      Wrigley, Stuart  and
      Song, Xingyi  and
      Gow-Smith, Edward  and
      Bawden, Rachel  and
      S{\'a}nchez-Cartagena, V{\'i}ctor M  and
      Cadwell, Patrick  and
      Lapshinova-Koltunski, Ekaterina  and
      Cabarr{\~a}o, Vera  and
      Chatzitheodorou, Konstantinos  and
      Nurminen, Mary  and
      Kanojia, Diptesh  and
      Moniz, Helena",
    booktitle = "Proceedings of the 25th Annual Conference of the European Association for Machine Translation (Volume 1)",
    month = jun,
    year = "2024",
    address = "Sheffield, UK",
    publisher = "European Association for Machine Translation (EAMT)",
    url = "https://aclanthology.org/2024.eamt-1.10/",
    pages = "68--79"
}

@misc{sellam2020bleurtlearningrobustmetrics,
      title={BLEURT: Learning Robust Metrics for Text Generation}, 
      author={Thibault Sellam and Dipanjan Das and Ankur P. Parikh},
      year={2020},
      eprint={2004.04696},
      archivePrefix={arXiv},
      primaryClass={cs.CL},
      url={https://arxiv.org/abs/2004.04696}, 
}

@inproceedings{ruiter-etal-2019-self,
    title = "Self-Supervised Neural Machine Translation",
    author = "Ruiter, Dana  and
      Espa{\~n}a-Bonet, Cristina  and
      van Genabith, Josef",
    editor = "Korhonen, Anna  and
      Traum, David  and
      M{\`a}rquez, Llu{\'i}s",
    booktitle = "Proceedings of the 57th Annual Meeting of the Association for Computational Linguistics",
    month = jul,
    year = "2019",
    address = "Florence, Italy",
    publisher = "Association for Computational Linguistics",
    url = "https://aclanthology.org/P19-1178/",
    doi = "10.18653/v1/P19-1178",
    pages = "1828--1834"
}

@misc{schulman2017proximalpolicyoptimizationalgorithms,
      title={Proximal Policy Optimization Algorithms}, 
      author={John Schulman and Filip Wolski and Prafulla Dhariwal and Alec Radford and Oleg Klimov},
      year={2017},
      eprint={1707.06347},
      archivePrefix={arXiv},
      primaryClass={cs.LG},
      url={https://arxiv.org/abs/1707.06347}, 
}

@inproceedings{popovic2015chrf,
  title={chrF: character n-gram F-score for automatic MT evaluation},
  author={Popovi{\'c}, Maja},
  booktitle={Proceedings of the tenth workshop on statistical machine translation},
  pages={392--395},
  year={2015}
}

@inproceedings{papineni-etal-2002-bleu,
    title = "{B}leu: a Method for Automatic Evaluation of Machine Translation",
    author = "Papineni, Kishore  and
      Roukos, Salim  and
      Ward, Todd  and
      Zhu, Wei-Jing",
    editor = "Isabelle, Pierre  and
      Charniak, Eugene  and
      Lin, Dekang",
    booktitle = "Proceedings of the 40th Annual Meeting of the Association for Computational Linguistics",
    month = jul,
    year = "2002",
    address = "Philadelphia, Pennsylvania, USA",
    publisher = "Association for Computational Linguistics",
    url = "https://aclanthology.org/P02-1040/",
    doi = "10.3115/1073083.1073135",
    pages = "311--318"
}

@inproceedings{koehn2017six,
  title={Six Challenges for Neural Machine Translation},
  author={Koehn, Philipp and Knowles, Rebecca},
  booktitle={Proceedings of the First Workshop on Neural Machine Translation},
  year={2017}
}

@inproceedings{ev,
  title={Two New Evaluation Datasets for Low-Resource Machine Translation: Nepali-English and Sinhala-English},
  author={Guzman, Francisco and Chen, Peng-Jen and Ott, Myle and Pino, Juan and Lample, Guillaume and Koehn, Philipp and Chaudhary, Vishrav and Ranzato, Marc'Aurelio},
  journal={arXiv preprint arXiv:1902.01382},
  year={2019},
  booktitle={nvm}
}

@Inbook{Fisher1992,
author="Fisher, R. A.",
title="Statistical Methods for Research Workers",
bookTitle="Breakthroughs in Statistics: Methodology and Distribution",
year="1992",
publisher="Springer New York",
address="New York, NY",
pages="66--70",
isbn="978-1-4612-4380-9",
doi="10.1007/978-1-4612-4380-9_6",
url="https://doi.org/10.1007/978-1-4612-4380-9_6"
}

@inproceedings{mcnamee2023extensive,
  title={An extensive exploration of back-translation in 60 languages},
  author={McNamee, Paul and Duh, Kevin},
  booktitle={Findings of the Association for Computational Linguistics: ACL 2023},
  pages={8166--8183},
  year={2023}
}

@article{hendy2023good,
  title={How good are gpt models at machine translation? a comprehensive evaluation},
  author={Hendy, Amr and Abdelrehim, Mohamed and Sharaf, Amr and Raunak, Vikas and Gabr, Mohamed and Matsushita, Hitokazu and Kim, Young Jin and Afify, Mohamed and Awadalla, Hany Hassan},
  journal={arXiv preprint arXiv:2302.09210},
  year={2023}
}

@article{haddow-etal-2022-survey,
    title = "Survey of Low-Resource Machine Translation",
    author = "Haddow, Barry  and
      Bawden, Rachel  and
      Miceli Barone, Antonio Valerio  and
      Helcl, Jind{\v{r}}ich  and
      Birch, Alexandra",
    journal = "Computational Linguistics",
    volume = "48",
    number = "3",
    month = sep,
    year = "2022",
    address = "Cambridge, MA",
    publisher = "MIT Press",
    url = "https://aclanthology.org/2022.cl-3.6/",
    doi = "10.1162/coli_a_00446",
    pages = "673--732"
}

@inproceedings{conneau-etal-2020-unsupervised,
    title = "Unsupervised Cross-lingual Representation Learning at Scale",
    author = "Conneau, Alexis  and
      Khandelwal, Kartikay  and
      Goyal, Naman  and
      Chaudhary, Vishrav  and
      Wenzek, Guillaume  and
      Guzm{\'a}n, Francisco  and
      Grave, Edouard  and
      Ott, Myle  and
      Zettlemoyer, Luke  and
      Stoyanov, Veselin",
    editor = "Jurafsky, Dan  and
      Chai, Joyce  and
      Schluter, Natalie  and
      Tetreault, Joel",
    booktitle = "Proceedings of the 58th Annual Meeting of the Association for Computational Linguistics",
    month = jul,
    year = "2020",
    address = "Online",
    publisher = "Association for Computational Linguistics",
    url = "https://aclanthology.org/2020.acl-main.747/",
    doi = "10.18653/v1/2020.acl-main.747",
    pages = "8440--8451"
}

@misc{nllbteam2022languageleftbehindscaling,
      title={No Language Left Behind: Scaling Human-Centered Machine Translation}, 
      author={NLLB Team and Marta R. Costa-jussà and James Cross and Onur Çelebi and Maha Elbayad and Kenneth Heafield and Kevin Heffernan and Elahe Kalbassi and Janice Lam and Daniel Licht and Jean Maillard and Anna Sun and Skyler Wang and Guillaume Wenzek and Al Youngblood and Bapi Akula and Loic Barrault and Gabriel Mejia Gonzalez and Prangthip Hansanti and John Hoffman and Semarley Jarrett and Kaushik Ram Sadagopan and Dirk Rowe and Shannon Spruit and Chau Tran and Pierre Andrews and Necip Fazil Ayan and Shruti Bhosale and Sergey Edunov and Angela Fan and Cynthia Gao and Vedanuj Goswami and Francisco Guzmán and Philipp Koehn and Alexandre Mourachko and Christophe Ropers and Safiyyah Saleem and Holger Schwenk and Jeff Wang},
      year={2022},
      eprint={2207.04672},
      archivePrefix={arXiv},
      primaryClass={cs.CL},
      url={https://arxiv.org/abs/2207.04672}, 
}

@misc{shao2024deepseekmathpushinglimitsmathematical,
      title={DeepSeekMath: Pushing the Limits of Mathematical Reasoning in Open Language Models}, 
      author={Zhihong Shao and Peiyi Wang and Qihao Zhu and Runxin Xu and Junxiao Song and Xiao Bi and Haowei Zhang and Mingchuan Zhang and Y. K. Li and Y. Wu and Daya Guo},
      year={2024},
      eprint={2402.03300},
      archivePrefix={arXiv},
      primaryClass={cs.CL},
      url={https://arxiv.org/abs/2402.03300}, 
}

@misc{sennrich2016improvingneuralmachinetranslation,
      title={Improving Neural Machine Translation Models with Monolingual Data}, 
      author={Rico Sennrich and Barry Haddow and Alexandra Birch},
      year={2016},
      eprint={1511.06709},
      archivePrefix={arXiv},
      primaryClass={cs.CL},
      url={https://arxiv.org/abs/1511.06709}, 
}

@misc{feng2025mtr1zeroadvancingllmbasedmachine,
      title={MT-R1-Zero: Advancing LLM-based Machine Translation via R1-Zero-like Reinforcement Learning}, 
      author={Zhaopeng Feng and Shaosheng Cao and Jiahan Ren and Jiayuan Su and Ruizhe Chen and Yan Zhang and Zhe Xu and Yao Hu and Jian Wu and Zuozhu Liu},
      year={2025},
      eprint={2504.10160},
      archivePrefix={arXiv},
      primaryClass={cs.CL},
      url={https://arxiv.org/abs/2504.10160}, 
}

@misc{wu2018studyreinforcementlearningneural,
      title={A Study of Reinforcement Learning for Neural Machine Translation}, 
      author={Lijun Wu and Fei Tian and Tao Qin and Jianhuang Lai and Tie-Yan Liu},
      year={2018},
      eprint={1808.08866},
      archivePrefix={arXiv},
      primaryClass={cs.LG},
      url={https://arxiv.org/abs/1808.08866}, 
}

@misc{ranzato2016sequenceleveltrainingrecurrent,
      title={Sequence Level Training with Recurrent Neural Networks}, 
      author={Marc'Aurelio Ranzato and Sumit Chopra and Michael Auli and Wojciech Zaremba},
      year={2016},
      eprint={1511.06732},
      archivePrefix={arXiv},
      primaryClass={cs.LG},
      url={https://arxiv.org/abs/1511.06732}, 
}

@misc{maksai2018eliminatingexposurebiaslossevaluation,
      title={Eliminating Exposure Bias and Loss-Evaluation Mismatch in Multiple Object Tracking}, 
      author={Andrii Maksai and Pascal Fua},
      year={2018},
      eprint={1811.10984},
      archivePrefix={arXiv},
      primaryClass={cs.CV},
      url={https://arxiv.org/abs/1811.10984}, 
}

@book{Sutton1998,
  author = {Sutton, Richard S. and Barto, Andrew G.},
  edition = {Second},
  publisher = {The MIT Press},
  title = {Reinforcement Learning: An Introduction},
  url = {http://incompleteideas.net/book/the-book-2nd.html},
  year = {2018 }
}

@inproceedings{huang-1990-machine,
    title = "A Machine Translation System for the Target Language Inexpert",
    author = "Huang, Xiuming",
    booktitle = "{COLING} 1990 Volume 3: Papers presented to the 13th International Conference on Computational Linguistics",
    year = "1990",
    url = "https://aclanthology.org/C90-3074/"
}

@inproceedings{federmann-etal-2019-multilingual,
  title     = {Multilingual Whispers: Generating Paraphrases with Translation},
  author    = {Federmann, Christian and Elachqar, Oussama and Quirk, Chris},
  booktitle = {Proceedings of the 5th Workshop on Noisy User-generated Text (W-NUT 2019)},
  year      = {2019},
  address   = {Hong Kong, China},
  publisher = {Association for Computational Linguistics},
  pages     = {17--26},
  url       = {https://aclanthology.org/D19-5503/},
  doi       = {10.18653/v1/D19-5503}
}

@misc{lample2018unsupervisedmachinetranslationusing,
      title={Unsupervised Machine Translation Using Monolingual Corpora Only}, 
      author={Guillaume Lample and Alexis Conneau and Ludovic Denoyer and Marc'Aurelio Ranzato},
      year={2018},
      eprint={1711.00043},
      archivePrefix={arXiv},
      primaryClass={cs.CL},
      url={https://arxiv.org/abs/1711.00043}, 
}

@misc{bengio2015scheduled,
      title={Scheduled Sampling for Sequence Prediction with Recurrent Neural Networks}, 
      author={Samy Bengio and Oriol Vinyals and Navdeep Jaitly and Noam Shazeer},
      year={2015},
      eprint={1506.03099},
      archivePrefix={arXiv},
      primaryClass={cs.LG},
      url={https://arxiv.org/abs/1506.03099}, 
}

@article{chang-etal-2024-goldfish,
  title={Goldfish: Monolingual Language Models for 350 Languages},
  author={Chang, Tyler A. and Arnett, Catherine and Tu, Zhuowen and Bergen, Benjamin K.},
  journal={Preprint},
  year={2024},
  url={https://www.arxiv.org/abs/2408.10441},
}

@misc{rei2020cometneuralframeworkmt,
      title={COMET: A Neural Framework for MT Evaluation}, 
      author={Ricardo Rei and Craig Stewart and Ana C Farinha and Alon Lavie},
      year={2020},
      eprint={2009.09025},
      archivePrefix={arXiv},
      primaryClass={cs.CL},
      url={https://arxiv.org/abs/2009.09025}, 
}

@misc{paszke2019pytorchimperativestylehighperformance,
      title={PyTorch: An Imperative Style, High-Performance Deep Learning Library}, 
      author={Adam Paszke and Sam Gross and Francisco Massa and Adam Lerer and James Bradbury and Gregory Chanan and Trevor Killeen and Zeming Lin and Natalia Gimelshein and Luca Antiga and Alban Desmaison and Andreas Köpf and Edward Yang and Zach DeVito and Martin Raison and Alykhan Tejani and Sasank Chilamkurthy and Benoit Steiner and Lu Fang and Junjie Bai and Soumith Chintala},
      year={2019},
      eprint={1912.01703},
      archivePrefix={arXiv},
      primaryClass={cs.LG},
      url={https://arxiv.org/abs/1912.01703}, 
}

@inproceedings{wolf-etal-2020-transformers,
    title = "Transformers: State-of-the-Art Natural Language Processing",
    author = "Thomas Wolf and Lysandre Debut and Victor Sanh and Julien Chaumond and Clement Delangue and Anthony Moi and Pierric Cistac and Tim Rault and Rémi Louf and Morgan Funtowicz and Joe Davison and Sam Shleifer and Patrick von Platen and Clara Ma and Yacine Jernite and Julien Plu and Canwen Xu and Teven Le Scao and Sylvain Gugger and Mariama Drame and Quentin Lhoest and Alexander M. Rush",
    booktitle = "Proceedings of the 2020 Conference on Empirical Methods in Natural Language Processing: System Demonstrations",
    month = oct,
    year = "2020",
    address = "Online",
    publisher = "Association for Computational Linguistics",
    url = "https://www.aclweb.org/anthology/2020.emnlp-demos.6",
    pages = "38--45"
}

@misc{zhang2020bertscoreevaluatingtextgeneration,
      title={BERTScore: Evaluating Text Generation with BERT}, 
      author={Tianyi Zhang and Varsha Kishore and Felix Wu and Kilian Q. Weinberger and Yoav Artzi},
      year={2020},
      eprint={1904.09675},
      archivePrefix={arXiv},
      primaryClass={cs.CL},
      url={https://arxiv.org/abs/1904.09675}, 
}

@InProceedings{pmlr-v202-fernandes23a,
  title = 	 {Scaling Laws for Multilingual Neural Machine Translation},
  author =       {Fernandes, Patrick and Ghorbani, Behrooz and Garcia, Xavier and Freitag, Markus and Firat, Orhan},
  booktitle = 	 {Proceedings of the 40th International Conference on Machine Learning},
  pages = 	 {10053--10071},
  year = 	 {2023},
  editor = 	 {Krause, Andreas and Brunskill, Emma and Cho, Kyunghyun and Engelhardt, Barbara and Sabato, Sivan and Scarlett, Jonathan},
  volume = 	 {202},
  series = 	 {Proceedings of Machine Learning Research},
  month = 	 {23--29 Jul},
  publisher =    {PMLR},
  url = 	 {https://proceedings.mlr.press/v202/fernandes23a.html}
}

@misc{kaplan2020scalinglawsneurallanguage,
      title={Scaling Laws for Neural Language Models}, 
      author={Jared Kaplan and Sam McCandlish and Tom Henighan and Tom B. Brown and Benjamin Chess and Rewon Child and Scott Gray and Alec Radford and Jeffrey Wu and Dario Amodei},
      year={2020},
      eprint={2001.08361},
      archivePrefix={arXiv},
      primaryClass={cs.LG},
      url={https://arxiv.org/abs/2001.08361}, 
}

@misc{kingma2017adammethodstochasticoptimization,
      title={Adam: A Method for Stochastic Optimization}, 
      author={Diederik P. Kingma and Jimmy Ba},
      year={2017},
      eprint={1412.6980},
      archivePrefix={arXiv},
      primaryClass={cs.LG},
      url={https://arxiv.org/abs/1412.6980}, 
}

@misc{paulus2017deepreinforcedmodelabstractive,
      title={A Deep Reinforced Model for Abstractive Summarization}, 
      author={Romain Paulus and Caiming Xiong and Richard Socher},
      year={2017},
      eprint={1705.04304},
      archivePrefix={arXiv},
      primaryClass={cs.CL},
      url={https://arxiv.org/abs/1705.04304}, 
}

@InProceedings{10.1007/11941439_149,
author="van Zaanen, Menno
and Zwarts, Simon",
editor="Sattar, Abdul
and Kang, Byeong-ho",
title="Unsupervised Measurement of Translation Quality Using Multi-engine, Bi-directional Translation",
booktitle="AI 2006: Advances in Artificial Intelligence",
year="2006",
publisher="Springer Berlin Heidelberg",
address="Berlin, Heidelberg",
pages="1208--1214",
isbn="978-3-540-49788-2"
}

@inproceedings{lazaridou-etal-2020-multi,
    title = "Multi-agent Communication meets Natural Language: Synergies between Functional and Structural Language Learning",
    author = "Lazaridou, Angeliki  and
      Potapenko, Anna  and
      Tieleman, Olivier",
    editor = "Jurafsky, Dan  and
      Chai, Joyce  and
      Schluter, Natalie  and
      Tetreault, Joel",
    booktitle = "Proceedings of the 58th Annual Meeting of the Association for Computational Linguistics",
    month = jul,
    year = "2020",
    address = "Online",
    publisher = "Association for Computational Linguistics",
    url = "https://aclanthology.org/2020.acl-main.685/",
    doi = "10.18653/v1/2020.acl-main.685",
    pages = "7663--7674"
}

@inproceedings{NIPS2017_3f5ee243,
 author = {Vaswani, Ashish and Shazeer, Noam and Parmar, Niki and Uszkoreit, Jakob and Jones, Llion and Gomez, Aidan N and Kaiser, \L ukasz and Polosukhin, Illia},
 booktitle = {Advances in Neural Information Processing Systems},
 editor = {I. Guyon and U. Von Luxburg and S. Bengio and H. Wallach and R. Fergus and S. Vishwanathan and R. Garnett},
 pages = {},
 publisher = {Curran Associates, Inc.},
 title = {Attention is All you Need},
 url = {https://proceedings.neurips.cc/paper_files/paper/2017/file/3f5ee243547dee91fbd053c1c4a845aa-Paper.pdf},
 volume = {30},
 year = {2017}
}

@inproceedings{wang2020exposure,
  title={On Exposure Bias, Hallucination and Domain Shift in Neural Machine Translation},
  author={Wang, Weiyue and Sennrich, Rico},
  booktitle={Proceedings of the 58th Annual Meeting of the Association for Computational Linguistics},
  pages={3544--3552},
  year={2020}
}
\appendix

\section{Qualitative Examples}
\label{app:qualitative}
\subsection{Central Aymara}
\noindent
\textbf{Reference:} Ukax jichhax amuyt’añjamaxïwa, kipka irnaqawinakjamakiwa. Jaqinakax kunti munapki ch’amañchañawa. ¡kayak anatañ lurañ yant’äsma! Wali jasakiw apnaqañaxa, qutan samart’añjamaspawa jan ukasti qhuru umanakan chuyma jalsuyasirjamarakispawa.\\
\textbf{Source (EN):} Ah that makes a ton of sense, they seem like similar jobs at least in some respects. Encouraging people and helping them meet their goals. You should totally try kayaking! It's super versatile, it can be relaxing on a lake or thrilling doing whitewater.\\
\textbf{Vanilla (system output):} Ah, al à sens, a somein lavoris similârs almancul in cualchi sens. incoragjâ lis personis e judâlis a rivâ ai lôr obietîfs. Tu varessis di provâ il kayak! al è une vore versatile, al pues jessi un lavôr di\\
\textbf{Vanilla (EN gloss):} Ah, al à sens, a somin lavoris similarârs at least in some sens. incoragjâ lis personis e judâlis a rivâ ai lôr obietîfs. Tu varessis di provâ il kayak! al è une vore versatile, al pues jessi un lavôr di\\
\textbf{Trained (system output):} Ah, al à sens, a somein lavoris similis almancul in cualchi sens. Incoragjâ lis personis e judâlis a rivâ ai lôr obietîfs. Tu varessis di provâ a fâ il kayak! Al è une vore versatile, al pues jessi ri\\
\textbf{Trained (EN gloss):} Ah, in a sense, some people work similarly at least in some senses. Incoragjâ the people and Jews to achieve their objectives. You have a variety of ways to try to do the kayaking! Al è une vore versatile, you can be very versatile, and you can be the best in the world.

\vspace{0.8em}
\noindent
\textbf{Reference:} O ai apene sfornât i biscos PB cu la forme dai mei cjans!\\
\textbf{Source (EN):} I baked PB cookies in the shape of my dogs!\\
\textbf{Vanilla (system output):} O ai fat biscots di PB a forme dai miei cans!\\
\textbf{Vanilla (EN gloss):} Or PB biscuits shaped like my dogs!\\
\textbf{Trained (system output):} O ài fat biscuts PB a forme dai miei cans!\\
\textbf{Trained (EN gloss):} I made PB cookies in the shape of my dogs!

\vspace{0.8em}
\noindent
\textbf{Reference:} O soi sigur che al divente mat, ma tu asreson, il mangja co cusini no l’è just par lôr. Perciò o ten simpri sot man al di bon par lôr quant co cusini.\\
\textbf{Source (EN):} I am sure the smells drive him crazy, but you are right. The food I bake is not good for them. I have dog treats on hand anytime I bake for this reason.\\
\textbf{Vanilla (system output):} O soi sigure che i odôrs lu fasin deventâ mat, ma tu âs reson. Il mangjâ che o fâs no je buine par lôr. Par chest motîf o ai simpri lis cjasis a man.\\
\textbf{Vanilla (EN gloss):} I'm sure the smells drive them crazy, but you're right. The food I make isn't good for them. That's why I always keep the houses handy.\\
\textbf{Trained (system output):} O soi sigure che i odôrs lu fasin deventâ mat, ma tu âs reson. Il mangjâ che o fâs jo nol è bon par lôr. Par cheste reson o ai simpri lis cjampanis pai cans in man.\\
\textbf{Trained (EN gloss):} I am sure the smells make them go crazy, but you are right. The food I make is not good for them. That is why I always have the dogs' bells in my hand.

\subsection{Friulian}
\noindent
\textbf{Reference:} O ai apene sfornât i biscos PB cu la forme dai mei cjans!\\
\textbf{Source (EN):} I baked PB cookies in the shape of my dogs!\\
\textbf{Vanilla (system output):} O ai fat biscots di PB a forme dai miei cans!\\
\textbf{Vanilla (EN gloss):} Or PB biscuits shaped like my dogs!\\
\textbf{Trained (system output):} O ài fat biscuts PB a forme dai miei cans!\\
\textbf{Trained (EN gloss):} I made PB cookies in the shape of my dogs!

\vspace{0.8em}
\noindent
\textbf{Reference:} O soi sigur che al divente mat, ma tu asreson, il mangja co cusini no l’è just par lôr. Perciò o ten simpri sot man al di bon par lôr quant co cusini.\\
\textbf{Source (EN):} I am sure the smells drive him crazy, but you are right. The food I bake is not good for them. I have dog treats on hand anytime I bake for this reason.\\
\textbf{Vanilla (system output):} O soi sigure che i odôrs lu fasin deventâ mat, ma tu âs reson. Il mangjâ che o fâs no je buine par lôr. Par chest motîf o ai simpri lis cjasis a man.\\
\textbf{Vanilla (EN gloss):} I'm sure the smells drive them crazy, but you're right. The food I make isn't good for them. That's why I always keep the houses handy.\\
\textbf{Trained (system output):} O soi sigure che i odôrs lu fasin deventâ mat, ma tu âs reson. Il mangjâ che o fâs jo nol è bon par lôr. Par cheste reson o ai simpri lis cjampanis pai cans in man.\\
\textbf{Trained (EN gloss):} I am sure the smells make them go crazy, but you are right. The food I make is not good for them. That is why I always have the dogs' bells in my hand.

\vspace{0.8em}
\noindent
\textbf{Reference:} No tant. Stât impegnât a ciri plas gnûs. Soi sigû co tacarai a vore a mae quat co sarai stac di cusinà.\\
\textbf{Source (EN):} Not recently. I have been consumed by trying to find recipes. I tend to hobby jump quite a bit. I am sure I will start knitting like crazy once I get tired of baking.\\
\textbf{Vanilla (system output):} No dome di pôc, mi soi consumade a cirî ricetis, o soi propit un pôc a saltâ fûr di hobby, o soi sigure che o fasarai il tricot come une matete cuant che mi stufarà di fâ il pan.\\
\textbf{Vanilla (EN gloss):} Not only that, I've been consumed with finding recipes, or I've been kind of jumping out of my hobby, or I'm sure I'll be knitting like a mat when I get tired of baking.\\
\textbf{Trained (system output):} No di pôc timp. O soi stât consumât a cirî ricetis. O ten a saltâ un grum di hobbies. O soi sigûr che o tacarà a fâ il tricot come un mat cuant che o varès stufât di fâ il pan.\\
\textbf{Trained (EN gloss):} Not recently. I have been consumed looking for recipes. I want to skip a lot of hobbies. I am sure that I will start knitting like crazy when I get tired of baking bread.

\begin{figure*}[t]
    \centering

    \begin{subcaptionblock}{0.24\textwidth}
        \centering
        \includegraphics[width=\linewidth]{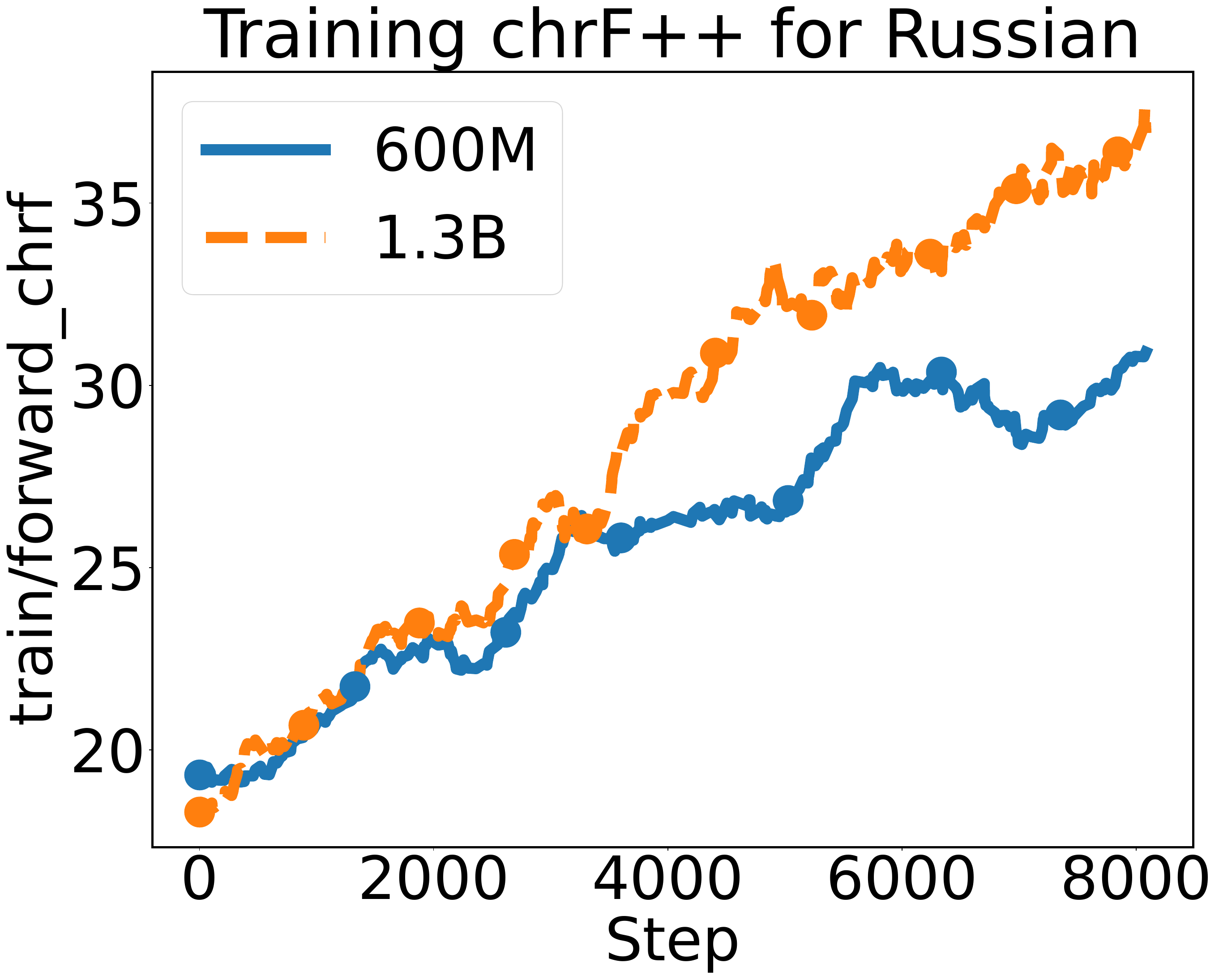}
    \end{subcaptionblock}\hfill
    \begin{subcaptionblock}{0.24\textwidth}
        \centering
        \includegraphics[width=\linewidth]{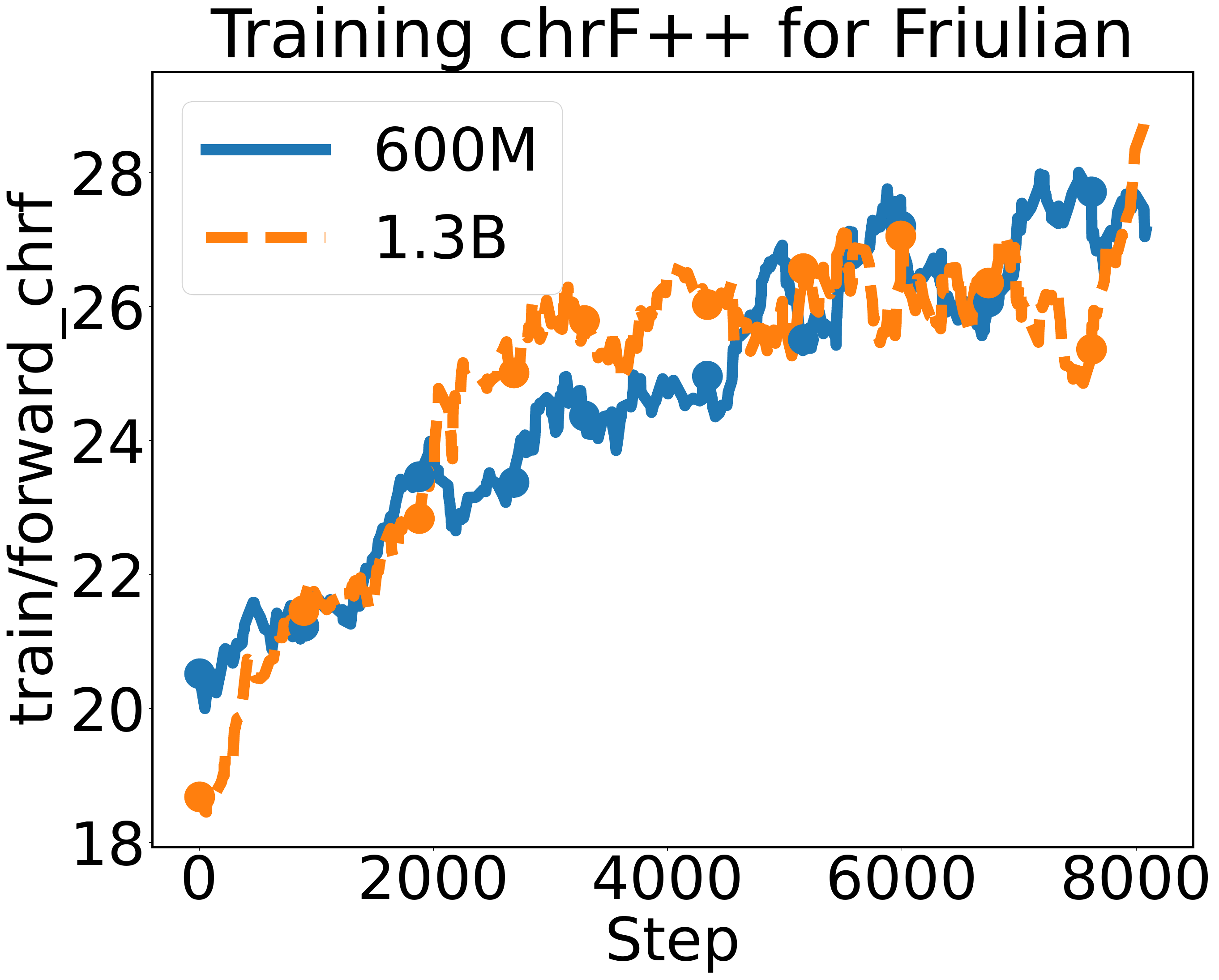}
    \end{subcaptionblock}\hfill
    \begin{subcaptionblock}{0.24\textwidth}
        \centering
        \includegraphics[width=\linewidth]{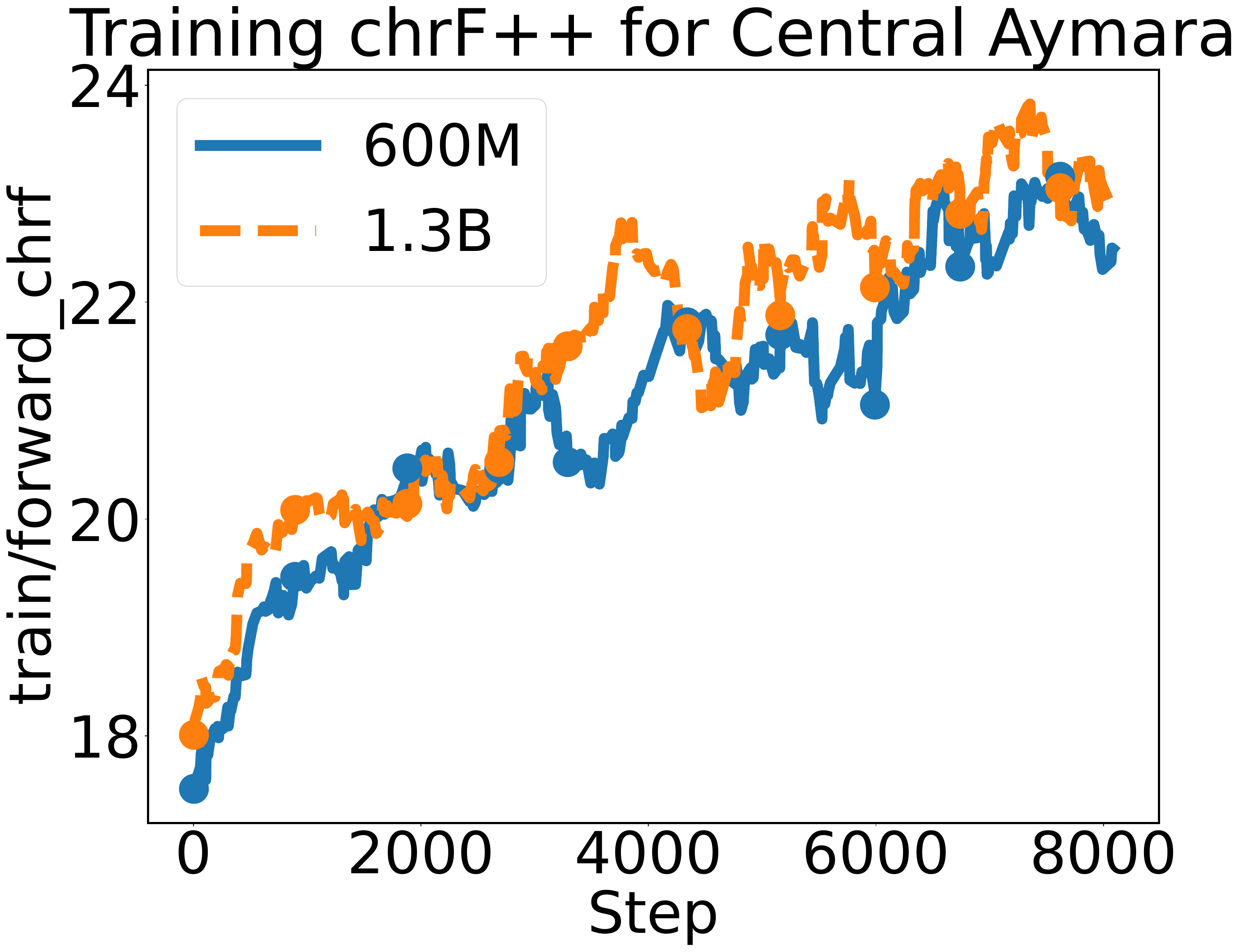}
    \end{subcaptionblock}\hfill
    \begin{subcaptionblock}{0.24\textwidth}
        \centering
        \includegraphics[width=\linewidth]{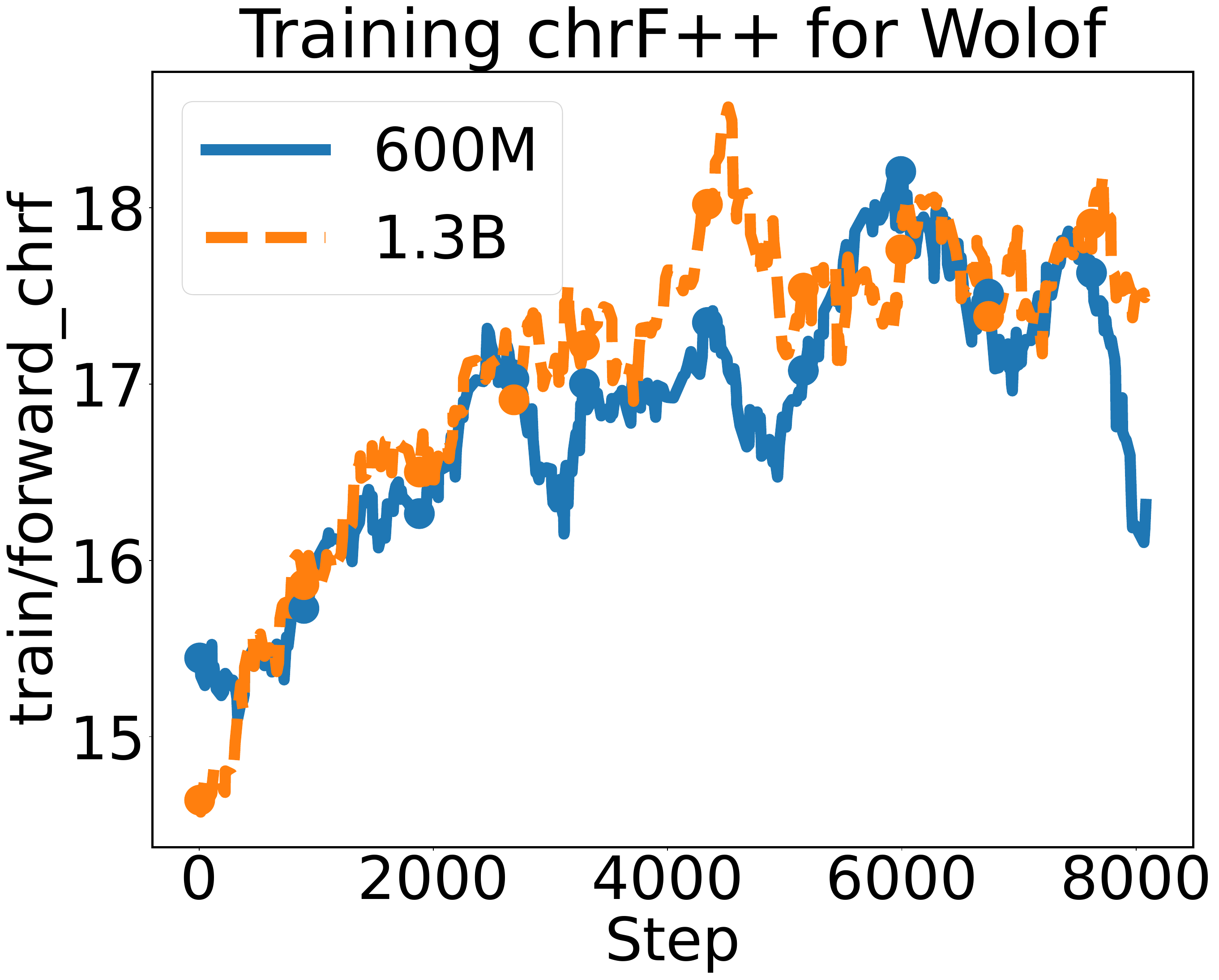}
    \end{subcaptionblock}

    \vspace{0.8em}

    \begin{subcaptionblock}{0.24\textwidth}
        \centering
        \includegraphics[width=\linewidth]{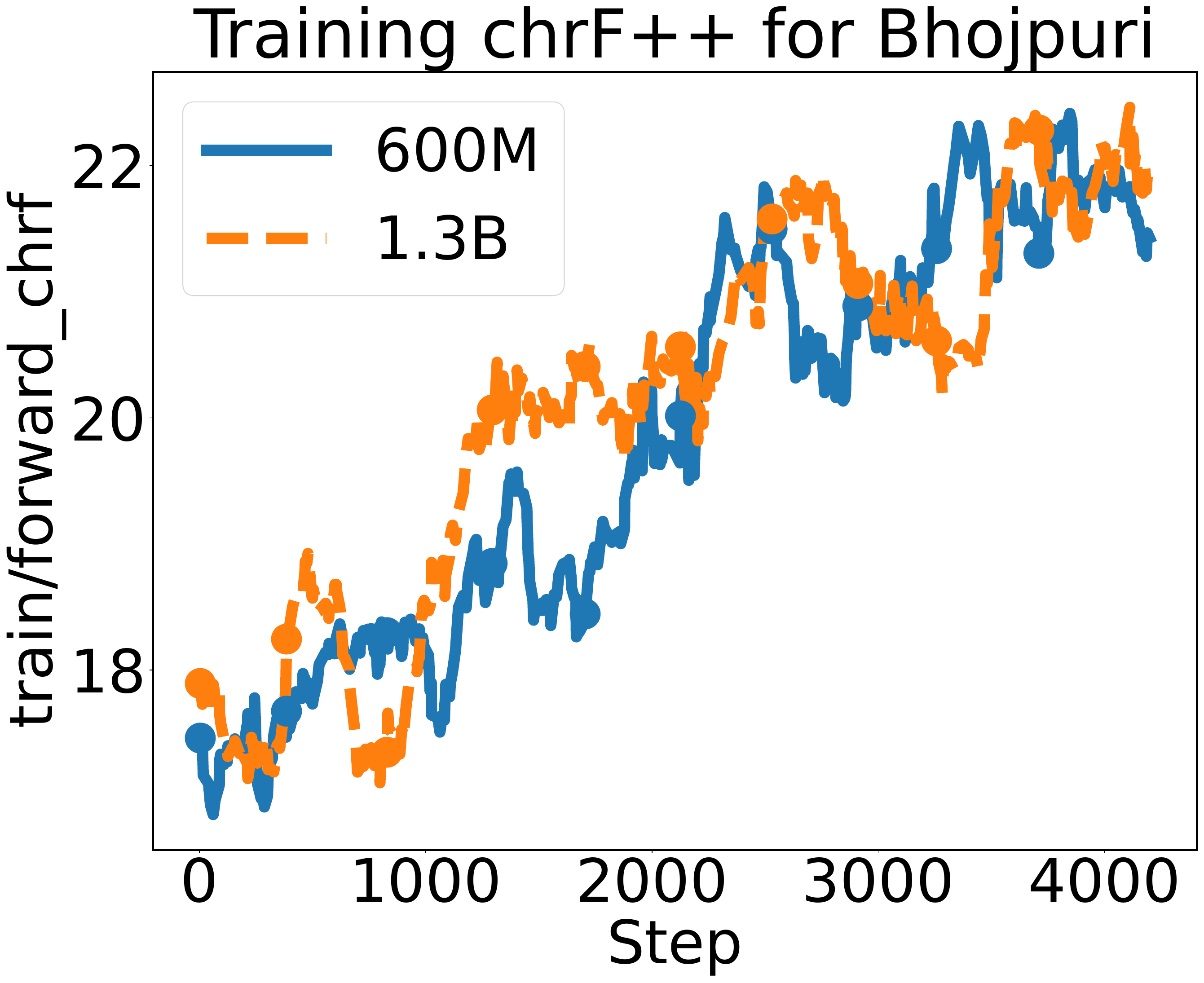}
    \end{subcaptionblock}
    \begin{subcaptionblock}{0.24\textwidth}
        \centering
        \includegraphics[width=\linewidth]{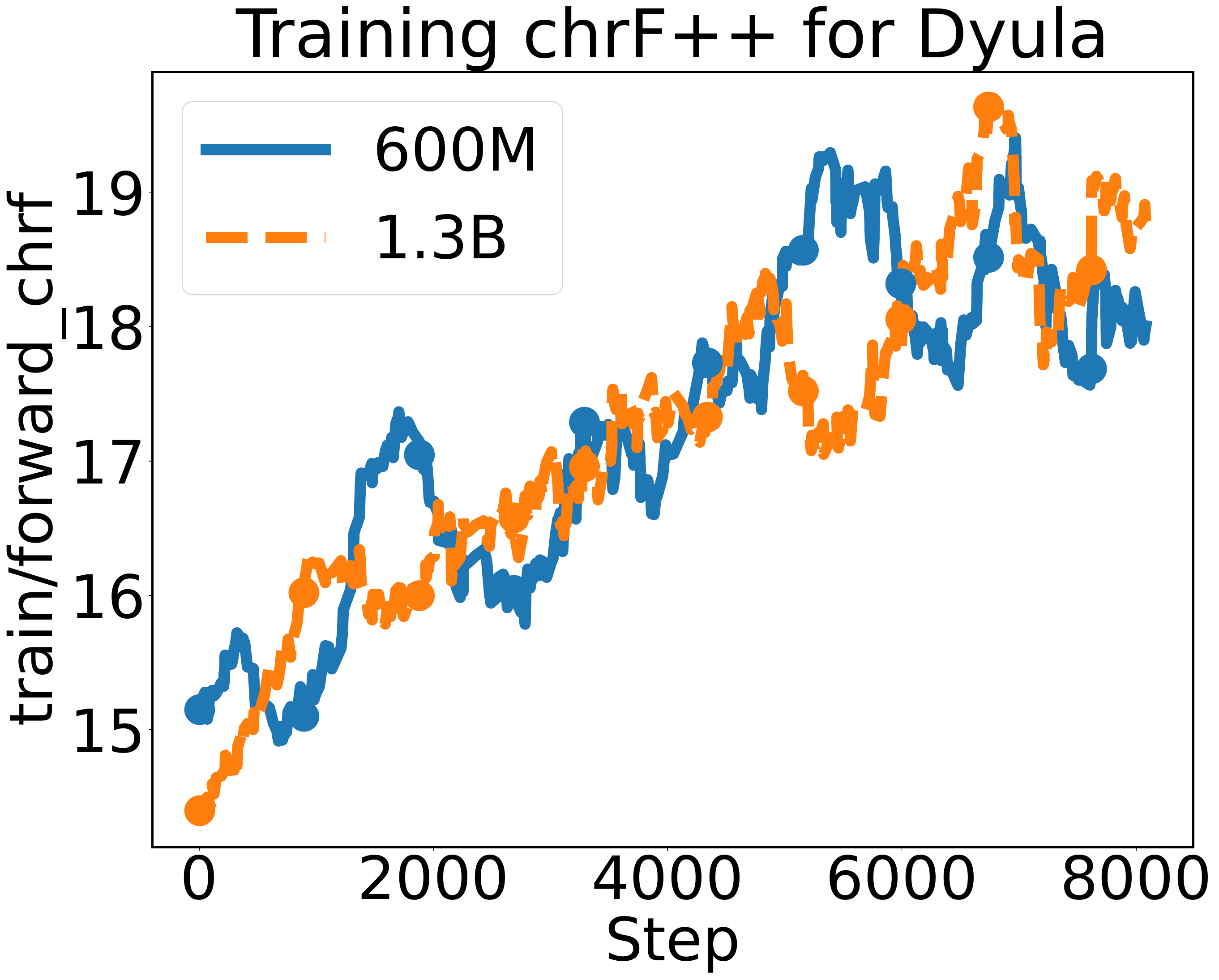}
    \end{subcaptionblock}
    \caption{ChrF++ reward training curves after 8K optimization steps for the 6 languages for both model sizes (600M and 1.3B).}
    \label{fig:app-forward-backward-curves}
\end{figure*}
\subsection{Russian}
\noindent
\textbf{Reference:} Da, no ya podumyvayu provesti vikend v etom kan'yone ryadom so mnoy. Eto ne plyazh, no i ne les, dolzhno byt' zdorovo.\\
\textbf{Source (EN):} I do, but I am thinking of doing a weekend in this canyon near me. It's not the beach, but it's nto a forest either, should be cool.\\
\textbf{Vanilla (system output):} Ya dumayu provesti vykhodnye v kan'one nedaleko ot menya.\\
\textbf{Vanilla (EN gloss):} I'm thinking of spending the weekend in a canyon near me.\\
\textbf{Trained (system output):} Da, no ya dumayu provesti vykhodnye v etom kan'one ryadom so mnoy. Eto ne plyazh, no eto tozhe les, dolzhno byt' kruto.\\
\textbf{Trained (EN gloss):} Yes, but I'm thinking of spending the weekend in that canyon next to me. It's not the beach, but it's also the forest, so that should be cool.

\vspace{0.8em}
\noindent
\textbf{Reference:} Tochno. Kak tvoya sem'ya? Skol'ko vas?\\
\textbf{Source (EN):} That's right. How is your family? how many of you are there?\\
\textbf{Vanilla (system output):} Kak vasha sem'ya?\\
\textbf{Vanilla (EN gloss):} How's your family?\\
\textbf{Trained (system output):} Tochno. Kak vasha sem'ya? Skol'ko vas tam?\\
\textbf{Trained (EN gloss):} Exactly. How is your family? How many are you there?

\vspace{0.8em}
\noindent
\textbf{Reference:} Nikakikh problem! Skol'ko i kakikh? 
\textbf{Source (EN):} No problem at all! How many and which ones?\\
\textbf{Vanilla (system output):} -- Skol'ko i kakie?\\
\textbf{Vanilla (EN gloss):} - How many and what?\\
\textbf{Trained (system output):} Nikakikh problem! Skol'ko i kakie?\\
\textbf{Trained (EN gloss):} No problem! How many and what?

\subsection{Wolof}
\noindent
\textbf{Reference:} Kuréelu réewum Suède bu wérgu-yaram (Rättsmedicinalverket) tàmbali woon na di def tests yi ci lu teel ci at mi.\\
\textbf{Source (EN):} Sweden's national Forensic Medicine Agency (Rättsmedicinalverket) started carrying out the tests earlier year.\\
\textbf{Vanilla (system output):} Bësum-bésum-bésum-bésum-bésum-bésum-bésum-bésum-bésum-bésum-bésum-bésum-bésum-bésum-bésum-bés\\
\textbf{Vanilla (EN gloss):} Day-day-day-day-day-day-day-day-day-day-day-day-day-day-day-day-day-day-day-day-day-day-day-day-day-day-day\\
\textbf{Trained (system output):} Bànqaasu Farãs gu réew mi ci Suweed (Rättsmedicinalverket) tàmbalee jëfandikoo ay seetlu ci njëlbéenug at mi.\\
\textbf{Trained (EN gloss):} The French national bank of Sweden (Rättsmedicinalverket) began using the studies at the beginning of the year.

\vspace{0.8em}
\noindent
\textbf{Reference:} Moom ak mbiipkat bu dëkk Bulgari Miroslav Petkov, dafay def ab konseer yu am bàkk bii\" Géej gu ñuul gi tasena ak Àll bu ñuul.\", wala sax, \" Balkan folklore\" daje naak ab njàng mu Allëmaañ bu kalasik ak taxawaay bu dëggër ci misiku brass.
\textbf{Source (EN):} Together with the Bulgarian trumpeter Miroslav Petkov, he is giving a concert that carries the slogan ``The Black Sea meets the Black Forest'', or even, ``Balkan folklore'' meets a classic German education with roots in brass music.\\
\textbf{Vanilla (system output):} Ci biir li muy ànd ak mbëjug buum buum buum buum buum buum buum buum buum buum buum buum buum buum buum buum buum buum buum buum buum buum buum buum buum buum\\
\textbf{Vanilla (EN gloss):} In addition to its use as an anti-inflammatory, it is also used as an anti-inflammatory and anti-inflammatory agent.\\
\textbf{Trained (system output):} Ci biir àndandoo ak mbëjug buulgar bi di Miroslav Petkov, mu ngi doon jox ab kaaraange bu am kàdduy waxtaan bi di ``Géej gu Màgg gi daje ak Géej gu Màgg gi'', walla sax, ``Balkan folkl\\
\textbf{Trained (EN gloss):} In collaboration with the Bulgarian writer Miroslav Petkov, she was developing a cover with the words of the speech ``The Old Sea meets the Great Sea'', or even, ``Balkan folklore''

\vspace{0.8em}
\noindent
\textbf{Reference:} Ñeen lañu! Amna sama benn makk bu goor ak samay waajur. Ñun goor yi deñoo bëgg bannexuloo ci benn sigaar ci ngoon gi.\\
\textbf{Source (EN):} There are four of us! I have an older brother and my two parents. Us men like to enjoy a cigar together in the evenings.\\
\textbf{Vanilla (system output):} Am nanu ñeenti nit, am naa magam ak samay maam ak sama baay.\\
\textbf{Vanilla (EN gloss):} We had four children, one of whom died in infancy.\\
\textbf{Trained (system output):} Am nanu ñeenti nit! Am naa mag bu góor ak samay ñaari waajur. Nun góor yi, bëgg nanu a lekk cigare ci ngoon ci ngoon.\\
\textbf{Trained (EN gloss):} We have four people! I have a older brother and my two parents. We men like to smoke in the afternoon in the afternoon.

\section{Additional Training Plots}
Figure~\ref{fig:app-forward-backward-curves} reports validation curves for the forward and backward translation models over 8K optimization steps.These curves are included for completeness and to illustrate the round-trip reinforcement learning training dynamics across languages. It's evident that there is no plateu and the models can further benefit from continued RL training.

\end{document}